%% file: example_arxiv.tex
\documentclass{article}
\usepackage[T1]{fontenc}
\usepackage[dvipsnames,table]{xcolor}
\definecolor{darkred}{RGB}{139,0,0}
\usepackage[preprint]{corl_2026} 
\usepackage{graphicx}
\usepackage{amsmath}
\usepackage{amssymb}
\usepackage{booktabs}
\usepackage{multirow}
\usepackage{graphicx}
\usepackage{wrapfig}
\usepackage{subcaption}
\definecolor{TableGray}{HTML}{F3F3F3}
\definecolor{BestCell}{HTML}{FCE8CC} 
\usepackage[most]{tcolorbox}
\tcbuselibrary{listings,breakable,skins}
\usepackage{listings}
\usepackage{tabularx}
\lstdefinestyle{promptstyle}{
  basicstyle=\ttfamily\scriptsize,
  breaklines=true,
  breakatwhitespace=true,
  breakautoindent=false,
  breakindent=0pt,
  postbreak={},
  columns=fullflexible,
  keepspaces=true,
  showstringspaces=false,
  frame=none,
  upquote=true,
  literate={_}{\textunderscore}1,
}

\newtcblisting[auto counter, number within=section]{promptbox}[2][]{
  listing only,
  listing options={style=promptstyle},
  colback=gray!4,
  colframe=gray!35,
  boxrule=0.4pt,
  arc=1mm,
  left=1mm,
  right=1mm,
  top=1mm,
  bottom=1mm,
  breakable,
  enhanced,
  title={Prompt~\thetcbcounter: #2},
  fonttitle=\bfseries\small,
  colbacktitle=darkred!85!black,
  #1
}

\newcommand{\best}[1]{\cellcolor{BestCell}\textbf{#1}}
\newcommand{\method}{\textbf{\textcolor{darkred}{OMG}}}
\usepackage{fontawesome5}
\usepackage[most]{tcolorbox}

\definecolor{LabelTextBG}{HTML}{E8F1FF}
\definecolor{LabelTextFG}{HTML}{1F5AA6}

\definecolor{LabelAudioBG}{HTML}{FFF1D6}
\definecolor{LabelAudioFG}{HTML}{B46A00}

\definecolor{LabelHumanBG}{HTML}{EAF7EA}
\definecolor{LabelHumanFG}{HTML}{2E7D32}

\newtcbox{\iconbox}[2][]{
    on line,
    boxsep=0pt,
    left=2.2pt,
    right=2.2pt,
    top=1.1pt,
    bottom=1.1pt,
    arc=2pt,
    boxrule=0pt,
    colback=#2,
    #1
}

\newcommand{\labeltext}{%
    \iconbox{LabelTextBG}{\textcolor{LabelTextFG}{\faIcon{keyboard}}}%
}

\newcommand{\labelaudio}{%
    \iconbox{LabelAudioBG}{\textcolor{LabelAudioFG}{\faIcon{assistive-listening-systems}}}%
}

\newcommand{\labelhuman}{%
    \iconbox{LabelHumanBG}{\textcolor{LabelHumanFG}{\faIcon{child}}}%
}

\newcommand{\ignore}[1]{}

\title{\method: \textcolor{darkred}{\underline{O}}mni-Modal \textcolor{darkred}{\underline{M}}otion \textcolor{darkred}{\underline{G}}eneration for Generalist Humanoid Control}

\author{
  {Siqiao Huang$^{*}$,
  Kun-Ying Lee$^{*}$,
  Dongming Qiao$^{*}$,
  Guanqi He$^{*}$,} \\
  \textbf{
  Zhenyu Wang,
  Yitang Li,
  Shaoting Zhu,
  Hang Zhao$^{\dagger}$
  }\\
  \text{Tsinghua University}\\
  $^{*}$Equal contribution \quad $^{\dagger}$Corresponding author \\
  \textbf{Project Page: \url{https://tsinghua-mars-lab.github.io/OMG/}}
}

\begin{document}
\maketitle

\vspace{-2em}
\begin{figure}[ht]
    \centering
    \includegraphics[width=\linewidth]{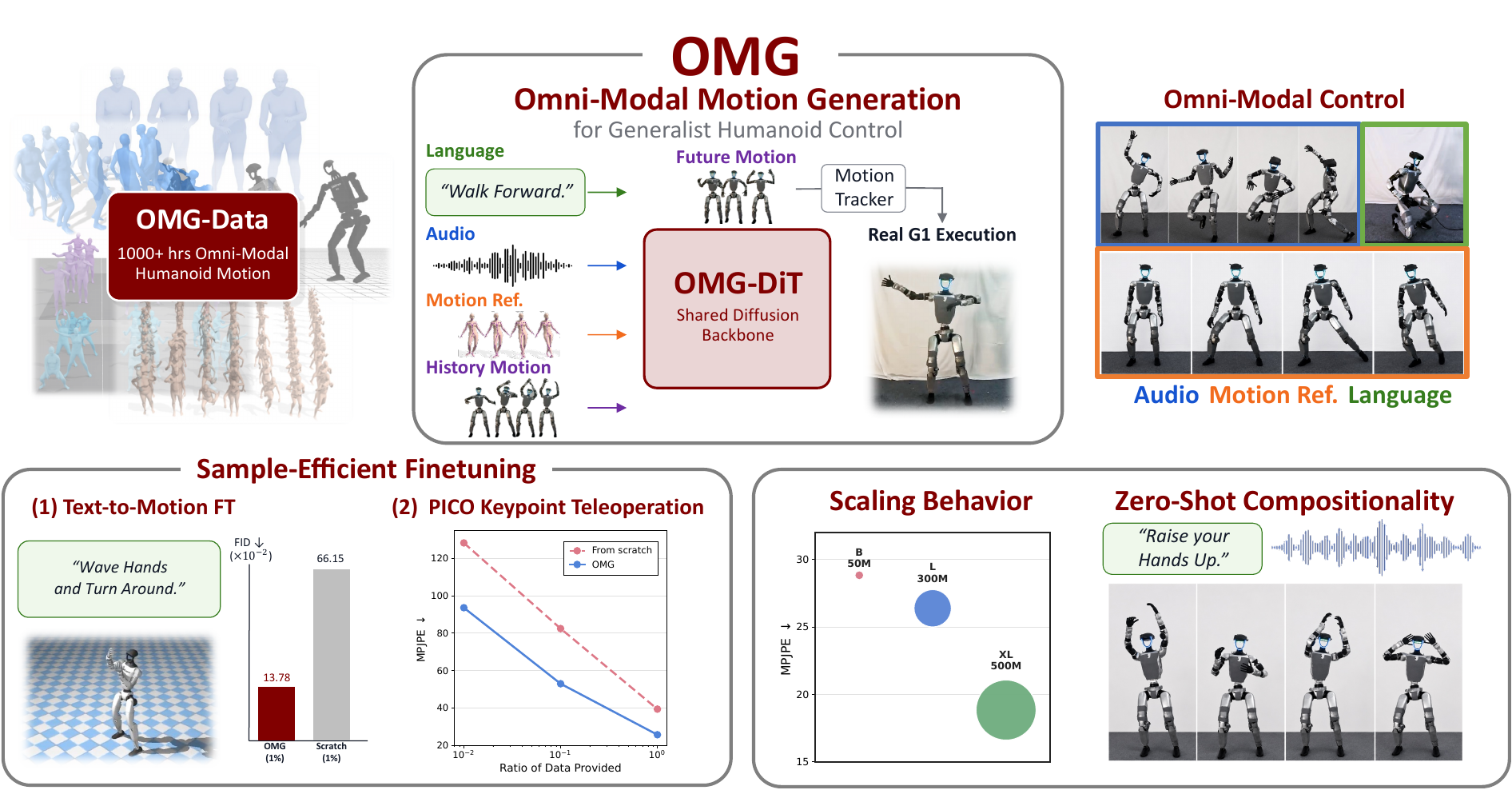}
    \caption{\textbf{Overview.} \method\ decomposes humanoid whole-body control into a scalable motion generation brain and a reactive motion tracking cerebellum. Built on \method\textbf{\textcolor{darkred}{-Data}}, a curation of 1000+ hours omni-modal humanoid motion data, \method\textbf{\textcolor{darkred}{-DiT}} maps language, audio, human reference, and their compositions into robot-executable future motions, which are deployed on a Unitree G1 in real time, paired with a pretrained motion tracker \cite{chen2026holomotion1}. This unified generator-tracker hierarchy enables general-purpose omni-modal control and sample-efficient adaptation to new tasks and modalities.}
    \label{fig:placeholder}
\end{figure}
\begin{abstract}
    Humanoid whole-body control has made significant progress in recent years, yet existing approaches remain limited to few-skill policies with heavy reward engineering, or motion trackers that are difficult to extend to new input modalities. We argue that the key to general-purpose humanoid control is to build a scalable \emph{brain}, a module capable of reasoning with diverse conditioning modalities, atop a reactive motion tracking \emph{cerebellum}, mirroring the hierarchical structure of biological motor systems. Two challenges arise in realizing this vision: acquiring a vast amount of high-quality data to achieve general purpose control, and equipping the generator with the capability to condition on compositional, extensible multi-modal inputs. We present \method, which addresses these challenges with a meticulous data curation, filtering and labeling pipeline, as well as a diffusion-based motion generation backbone that conditions on language, audio, and human reference motions. Extensive experiments validate \method\ as an omni-modal whole-body controller exhibiting state-of-the-art performance, model scaling behavior and efficient adaptation to new distributions and modalities, marking a concrete step toward foundation models for humanoid robots.
\end{abstract}


\input{sections/intro}

\input{sections/related}
\input{sections/data}
\input{sections/methods}

\input{sections/exp}
\input{sections/conclusion}
\input{sections/limitations}

\clearpage
\acknowledgments{The authors would like to sincerely thank Ziwen Zhuang, Zekun Qi, Haoyang Weng, and Yue Chen for their insightful discussions and feedback.}


\bibliography{example}  
\clearpage
\appendix

\input{app/theory}
\input{app/data_details}
\input{app/model_details}
\input{app/exp_details}
\input{app/ext_exp}
\end{document}

%% file: sections/intro.tex
\section{Introduction}

Humanoid whole-body control has made rapid progress,  enabling agile locomotion and dexterous loco-manipulation with reinforcement learning~\citep{zhuang2024humanoid,radosavovic2024learning,zhang2025falcon,li2025hold}. However, most existing policies remain tied to specific skills and reward designs, making them difficult to scale. Motion tracking provides a more scalable alternative by learning to follow reference motions~\citep{luo2025sonic,zhang2025track,liao2025beyondmimic}, but tracking alone largely replays given motions and offers limited autonomy under high-level, multi-modal human intent.

A promising abstraction is a generator-tracker hierarchy~\citep{serifi2024robot,tevet2025closd,xu2025parc,zhang2026learning,xie2026textop}, where an upstream motion generator translates high-level conditions into future whole-body trajectories, and a downstream tracker executes them on the robot~\citep{chen2026holomotion1,luo2025sonic}. Yet realizing this paradigm requires overcoming two key challenges. First, high-quality humanoid motion data is scarce, fragmented, and heterogeneous, unlike web-scale text or visual data~\citep{schuhmann2022laion,penedo2024fineweb}. Second, the motion generator must support diverse, composable, and extensible control modalities while remaining adaptable to new control interfaces.

To this end, we introduce \method, an \textcolor{darkred}{\underline{\textbf{O}}}mni-Modal \textcolor{darkred}{\underline{\textbf{M}}}otion \textcolor{darkred}{\underline{\textbf{G}}}eneration framework for generalist humanoid whole-body control. At the data level, we curate \textbf{\textcolor{darkred}{OMG-Data}}, a large-scale multi-modal humanoid motion corpus of \textcolor{darkred}{\textbf{1000+}} hours, by retargeting, filtering, annotating, and aligning heterogeneous motions into the Unitree G1 embodiment. At the model level, we instantiate \textbf{\textcolor{darkred}{OMG-DiT}}, a diffusion transformer backbone that maps language, audio, human-reference motions, and their compositions into robot-executable future trajectories. Importantly, new modalities can be incorporated through lightweight condition encoders while reusing the pretrained motion prior; unseen control signal combinations can be composed at inference  through guidance.

Extensive experiments demonstrate \method's\ capabilities in producing high-quality, physically executable motions across diverse modalities. Furthermore, \method\ exhibits foundation-model-like properties, including predictable scaling, sample-efficient adaptation, and zero-shot composition of control signals. These results suggest that generalist humanoid control can be advanced not only by stronger low-level controllers, but also by scaling the motion-generation brain that interfaces human intents with physical execution. In summary, our contributions are as follows:
\begin{itemize}
    \item We introduce \method, an omni-modal motion generation framework for generalist humanoid whole-body control, unifying diverse conditioning modalities under a shared backbone, mapping diverse human intents into physically executable robot trajectories.   
    \item We curate \textbf{\textcolor{darkred}{OMG-Data}}, a large-scale omni-modal humanoid motion corpus of \textcolor{darkred}{\textbf{1000+}} hours. Through a unified pipeline of retargeting, filtering, and annotation, we align motions into a unified motion space, making supervised scaling of humanoid motion generation possible.   
    \item We introduce \textbf{\textcolor{darkred}{OMG-DiT}}, a diffusion-based motion generation backbone that supports extensible and compositional conditioning from language, audio, and human reference motions, allowing new modalities to be incorporated through lightweight adaptation.    
    \item Through extensive experiments, we validate \method\ as an omni-modal whole-body controller, demonstrating foundation-model-like properties including model scaling behavior, few-shot adaptation, and zero-shot composition of control signals.
\end{itemize}

%% file: sections/related.tex
\section{Related Works}

\paragraph{Humanoid Whole-Body Control and Motion Tracking.}
Recent advances in reinforcement learning have greatly expanded the capability of humanoid whole-body controllers~\citep{zhuang2024humanoid,radosavovic2024learning,zhang2025falcon,li2025hold}. However, these systems are often specialized to particular task objectives and reward designs. Motion tracking~\citep{he2024learning,he2024omnih2o} offers a more scalable alternative, training a policy to execute human reference motions on the humanoid. 
Recent works~\citep{ze2025twist,luo2025sonic,chen2026holomotion1,zhang2025track,liao2025beyondmimic} scale this paradigm with broader motion corpora, yielding stronger controllers that robustly track out-of-domain reference motions. 
These trackers form a powerful low-level execution layer for humanoid control, yet they assume availability of reference motions or low-level commands at inference time, leaving open the upstream problem of generating robot-executable motions from high-level and multi-modal conditions.

\paragraph{Interactive and Multi-Modal Motion Generation.}
Motion generation~\citep{tevet2022human,zhang2022motiondiffuse} addresses the complementary problem of translating high-level conditions into motion references. In graphics, modern generative architectures~\citep{yang2023diffusion,peebles2023scalable} have substantially advanced motion generation systems, enabling expressive conditioning on text, audio, keypoints, and human references~\citep{liu2024emageunifiedholisticcospeech,zhang2025opendancemultimodalcontrollable3d,genmo2025,Kimodo2026}, while benefiting from scaling data~\citep{fan2025zerozeroshotmotiongeneration} and model parameters.
However, most such models remain human-motion generators that operate in an offline manner, neglecting the need for \textit{real-time interactive} control. 
Conversely, coupling motion generation with tracking systems has garnered increasing interest in the humanoid community~\citep{zhang2026learning,liao2025beyondmimic,tevet2025closd,serifi2024robot,xie2026textop}, yet these systems are typically limited to a single command modality or narrow task domain. This highlights an important missing intersection: general-purpose, omni-modal motion generation for real-time, robot-executable humanoid control.

%% file: sections/data.tex
\section{\textcolor{darkred}{OMG-Data}: Scaling Multi-Modal Data for Humanoid Motion Generation}
\begin{figure}
    \centering
    \includegraphics[width=\linewidth]{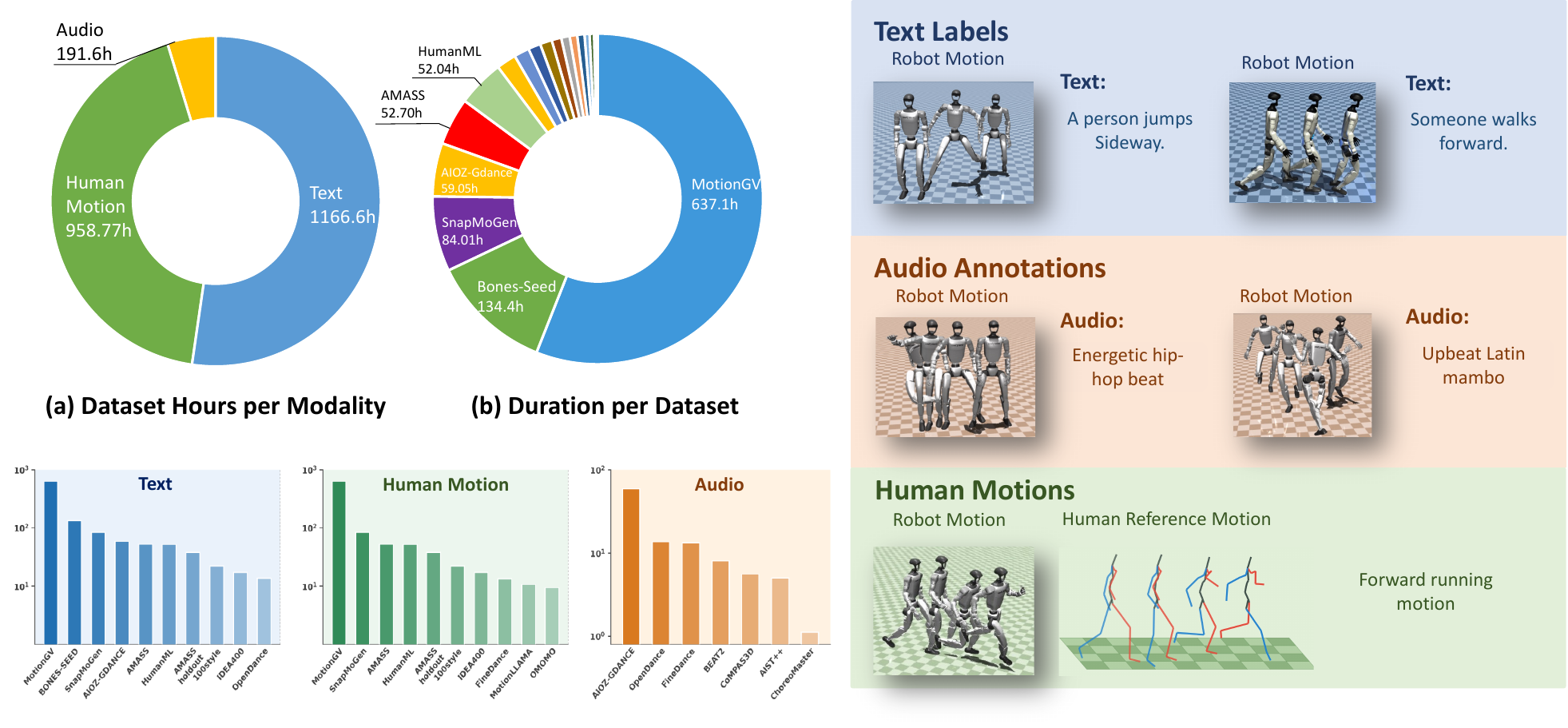}
    \caption{\textbf{Dataset Statistics of \textcolor{darkred}{OMG-Data}}. We curate a large-scale omni-modal humanoid motion corpus by aggregating heterogeneous datasets and unifying them into the Unitree G1 motion space.
Left: processed data statistics across conditioning modalities and source datasets.
Right: representative conditioning modalities, including language, audio, and human reference motions.}
    \label{fig:data_statistics}
\end{figure}
A central obstacle to scaling humanoid whole-body motion generation is the lack of a unified, robot-executable, and multi-modal motion corpus. Existing motion datasets are highly fragmented, and datasets differ substantially in their available label modality, motion quality, and annotation granularity. To address this challenge, we curate \textbf{\textcolor{darkred}{OMG-Data}}, a large-scale omni-modal humanoid motion corpus unified in the Unitree G1 embodiment of \textbf{1174.66} hours, realized through a carefully designed pipeline of curation, preprocessing, retargeting, annotating and filtering.

\paragraph{Data Curation and Preprocessing.} To establish a comprehensive corpus, we aggregate a diverse set of publicly available motion datasets across graphics and humanoid domains \cite{mahmood2019amass,harvey2020robust,li2023object,mason2022real,fan2025zerozeroshotmotiongeneration,snapmogen2025,li2023finedance,liu2024emageunifiedholisticcospeech,li2021learn,zhang2025motion,lin2023motionx,kim2025personabooth,Kimodo2026,choreomaster2021,burkanova2025salsanonverbalembodiedlanguage,aiozGdance,zhang2025opendancemultimodalcontrollable3d}. The consolidated dataset spans multiple modalities, encompassing text labels, audios, and human reference motions. Upon curation, the files first go through a validation check: all corrupted files, samples with broken links, and instances suffering from severe missing frames or invalid joint attributes are purged. For multi-modal sequence blocks that contain synchronized audio inputs (e.g., dance or co-speech gesture datasets like AIST++ \cite{li2021learn} and BEAT2 \cite{liu2024emageunifiedholisticcospeech}), we additionally preprocess the raw acoustic streams and audio feature vectors to be frame-aligned and temporally synchronized with corresponding motion clip trajectories.

\paragraph{Motion Retargeting and Annotation.} Since human motion representations vary across datasets, unifying the topology of motion representation becomes essential. Using General Motion Retargeting (GMR, \cite{ze2025gmr,ze2025twist,joao2025gmr}), we retarget all motion representations to the motion space of Unitree G1 humanoids. To enrich datasets that lack native language descriptions or require granular cross-modal semantics, we render the retargeted motion sequences in simulation and label them with \texttt{Seed-1.8}~\cite{seed2026seed1}, using multi-view rendered videos or representative keyframes as visual inputs. We then segment motions into training clips according to language annotation boundaries, audio phrase cuts, uniform windows, or sliding windows for long episodes.


\paragraph{Simulation-In-The-Loop Filtering.} To prevent kinematically invalid, self-colliding, or physically impossible joint configurations from contaminating the training mixture, we apply a filtering step via tracker execution in simulation. Concretely, given a sample, we roll out this motion sequence in the MuJoCo rigid-body simulator \cite{todorov2012mujoco} using a joint-space PD tracking controller. We design the fall detection heuristics as follows: Let $h_t$ denote root height and $\theta_t$ denote root tilt angle. We mark a frame as a fall frame if $h_t < 0.20$, $\theta_t > 85^\circ$, or $(h_t < 0.35 \land \theta_t > 60^\circ)$.
If the fall frame condition persists for over 10 consecutive frames, the trajectory is rejected. This physics-in-the-loop screening ensures that the processed dataset contains dynamically feasible and safe trajectories for humanoid deployment. Further details are provided in the Supplementary Material.

%% file: sections/methods.tex
\section{\textbf{\textcolor{darkred}{OMG-DiT}}: Unified Diffusion Backbone for Motion Generation}
\begin{figure}
    \centering
    \includegraphics[width=\linewidth]{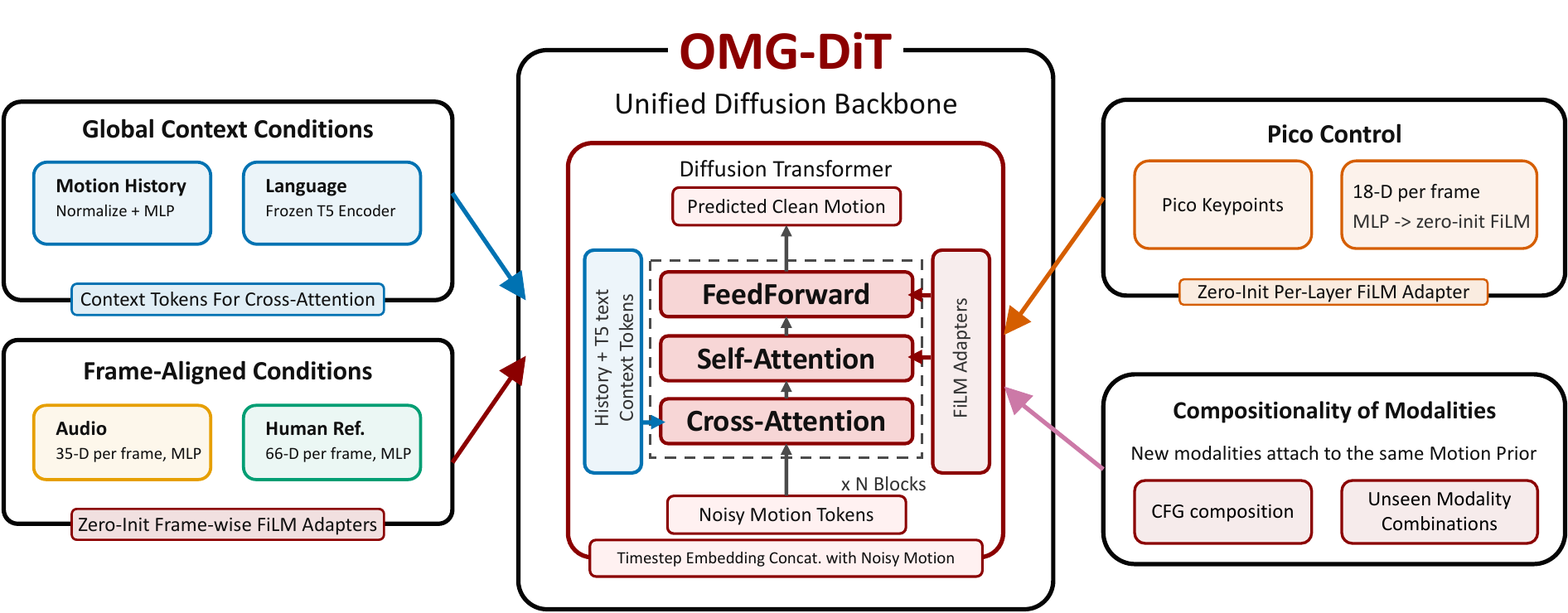}
    \caption{\method\textbf{\textcolor{darkred}{-DiT}} learns a shared diffusion backbone while enabling conditioning with modality-specific encoders. History motion and language are injected as global context tokens via cross-attention, whereas frame-aligned signals (i.e., audio and human reference motions) are injected through FiLM \cite{perez2018film} adapters. New modalities are attached non-invasively through zero-initialized adapters, and multiple conditions can be composed at inference via classifier-free guidance.}
    \label{fig:methods_main}
\end{figure}
Equipped with this large-scale multi-modal motion corpus, we instantiate \textbf{\textcolor{darkred}{OMG-DiT}}, a unified diffusion backbone that maps heterogeneous control modalities into robot-executable whole-body trajectories in real time. The key design principle is to decouple the \emph{motion prior} from the \emph{condition interface}: a shared denoising backbone maps the distribution of feasible motions, while modality-specific encoders translate high-level intents into steerable conditions over the shared manifold.

\subsection{Problem Formulation}
\label{sec:problem_formulation}
We consider the problem of controlling a humanoid robot to perform whole-body motions specified by multi-modal conditions. Given a set of conditioning variables
$\mathcal{C} = \{c^{(1)}, c^{(2)}, \ldots, c^{(N)}\}$, which may include language, audio and other modalities, our goal is to learn a policy $\pi$ that produces physically realizable whole-body actions $\mathbf{a}_t$ conditioned on $\mathcal{C}$ and history observations $\mathbf{o}_{\leq t}$. We decompose this policy learning problem into a cascaded generation-and-tracking formulation, or more commonly termed planning and inverse dynamics models in manipulation \cite{du2023learning,du2024video}:
\begin{equation*}
{
    \pi(\mathbf{a}_{t:t+H},\mathbf{o}_{t+1:t+H+1}\mid \mathbf{o}_{\leq t}, \mathcal{C})
    \approx
    \underbrace{
    \pi_\phi(\mathbf{o}_{t+1:t+H+1} \mid \mathbf{o}_{t-L:t},\mathcal{C})
    }_{\text{Motion Generation/Planning}}
    \cdot
    \underbrace{
    \pi_\psi(\mathbf{a}_{t:t+H}\mid \mathbf{o}_{t-L:t},\mathbf{o}_{t+1:t+H+1})
    }_{\text{Motion Tracking/IDM}} .
}
\end{equation*}
where $\pi_\phi$ is a high-level motion generator that predicts future whole-body reference observations from multi-modal conditions, and $\pi_\psi$ is a low-level motion tracker that converts these references into executable robot actions. In this work, we focus on motion generation, and leverage pre-existing general-purpose trackers (HoloMotion \cite{chen2026holomotion1}) as our low-level tracker.

\subsection{Diffusion Transformer Backbone}

\paragraph{Motion Representation.}
We represent each frame in a canonical root-centric coordinate frame defined by the last observed state $\mathbf{x}_t =
    [\tilde{\mathbf{p}}_t,\,
    \tilde{\mathbf{r}}_t,\,
    \boldsymbol{\theta}_t,\,
    \tilde{\mathbf{j}}_t] \in \mathbb{R}^{125}$,
where $\tilde{\mathbf{p}}_t$ and $\tilde{\mathbf{r}}_t$ denote the canonicalized root position and rotation, $\boldsymbol{\theta}_t$ denotes joint angles, and $\tilde{\mathbf{j}}_t$ denotes body-link positions. This root-centric representation removes global translation and heading ambiguity while preserving the full-body geometric structure required for motion generation and downstream tracking.

\paragraph{Model Architecture.} \textbf{\textcolor{darkred}{OMG-DiT}} is instantiated as a Diffusion Transformer (DiT \cite{peebles2023scalable}) backbone trained with an $\mathbf{x}$-prediction objective \cite{li2025back}. Drawing inspiration from recent advances in pixel-space diffusion \cite{li2025back,lu2026one}, we generate directly in motion space, avoiding training of motion encoders. History motion is injected via cross-attention, whereas the denoising timesteps are injected via channel-wise concatenating with the noisy motion features before input projection.

\paragraph{Training Objective.} We train \textbf{\textcolor{darkred}{OMG-DiT}} as a conditional diffusion model generating future motion trajectories $\mathbf{x}_{t+1:t+H}$ conditioned on recent history $\mathbf{x}_{t-L:t}$ and a subset of available conditioning modalities $\mathcal{C}_s \subseteq \mathcal{C}$. Given a diffusion timestep $\tau$, we corrupt the future segment as
\begin{equation}
    \mathbf{x}^{\tau}_{1:H}
    =
    \sqrt{\bar{\alpha}_{\tau}}\,\mathbf{x}_{1:H}
    +
    \sqrt{1-\bar{\alpha}_{\tau}}\,\boldsymbol{\epsilon},
    \qquad
    \boldsymbol{\epsilon} \sim \mathcal{N}(\mathbf{0},\mathbf{I}).
\end{equation}
To enable classifier-free guidance and improve robustness, we randomly drop each available conditioning modality during training with probability $p_{\mathrm{drop}}$. The resulting training objective is:
\begin{equation}
    \mathcal{L}
    =
    \mathbb{E}_{\mathbf{x},\mathcal{C}_s,\tau,\boldsymbol{\epsilon}}
    \left[
    \left\|
        \hat{\mathbf{x}}_{t+1:t+H}
        -
        \mathbf{x}_{t+1:t+H}
    \right\|_2^2
    \right], \text{where } \hat{\mathbf{x}}_{t+1:t+H}
    =
    \mathbf{x}_{\theta}
    \!\left(
        \mathbf{x}^{\tau}_{t+1:t+H},
        \tau,
        \mathbf{x}_{t-L:t},
        \mathcal{C}_s
    \right).
\end{equation}
This objective trains a single denoising backbone to model the feasible Unitree G1 motion manifold, while allowing different condition encoders to steer generation through a shared motion prior.

\subsection{Omni-Modal Condition Encoding}

\paragraph{Pretraining Condition Encoders.}
During pretraining, \textbf{\textcolor{darkred}{OMG-DiT}} supports conditioning on language, audio, and human-reference motion. Language annotations are encoded by a frozen T5 encoder \cite{raffel2020exploring} and injected into each DiT block through cross-attention, alongside encoded history frames. Frame-aligned modalities, including audio and human-reference motion, are projected through MLPs and injected into corresponding frames through per-layer FiLM modulation \cite{perez2018film}.

\paragraph{Task-specific Finetuning Encoders.}
To enable sample-efficient adaptation to downstream tasks with new modalities, we adopt non-invasive injection methods with zero-initialization. For example, while fine-tuning for Pico keypoint-conditioned teleoperation, Pico keypoints are represented as 18D frame-aligned features, and injected through zero-initialized FiLM \cite{perez2018film} adapters.

%% file: sections/exp.tex
\section{Experiments}
\begin{figure}
    \centering
    \includegraphics[width=0.95\linewidth, height=0.9\linewidth]{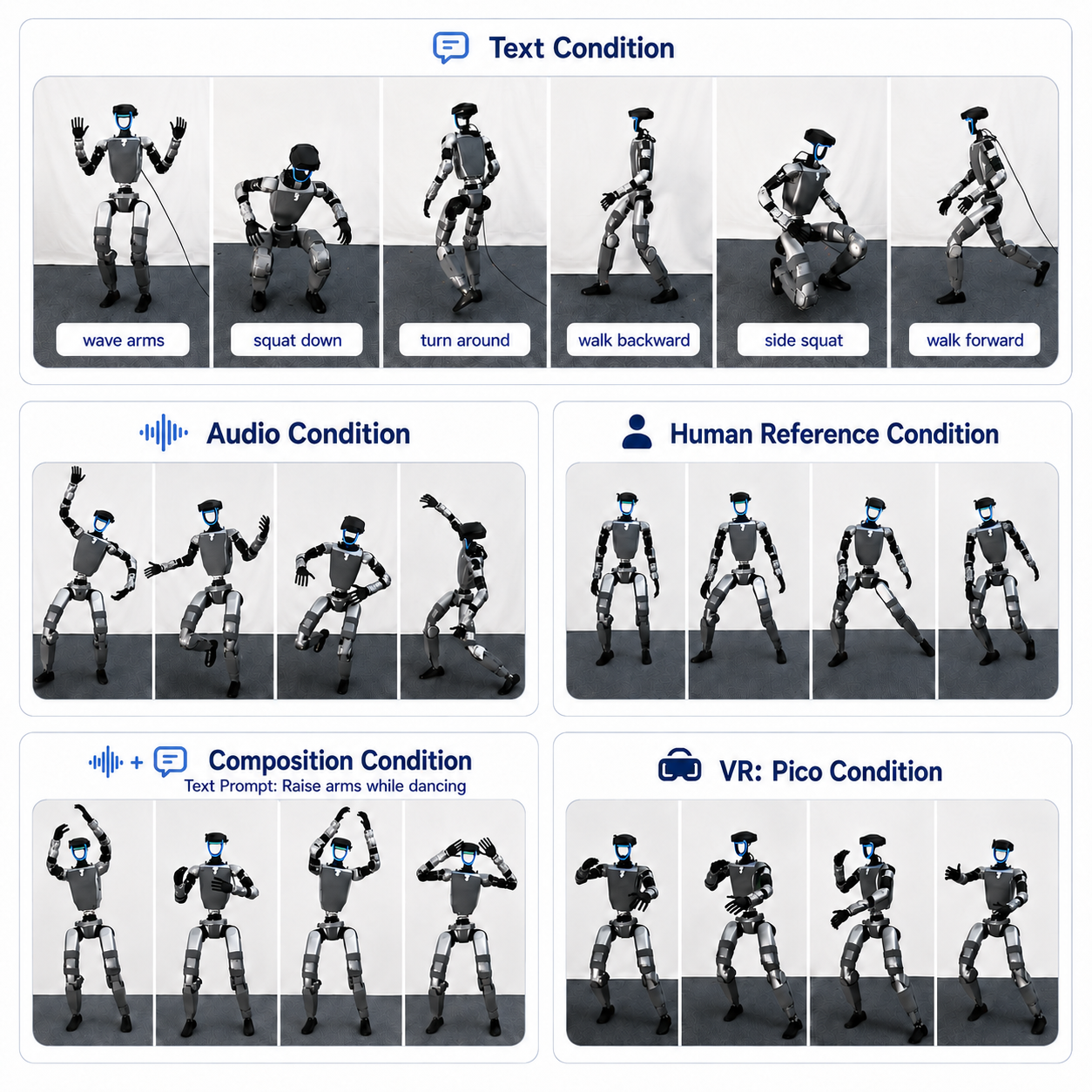}
    \vspace{-4mm}
\caption{\textbf{Real-World Omni-Modal Control.} \method\ generates diverse Unitree G1 motions across various conditioning modalities in real time, executable in the real world.}
    \label{fig:exp_main}
\end{figure}
Our experiments aim to answer the following questions: 
(1) How does \method\ compare against previous hierarchical humanoid control frameworks, such as graphics-based generation plus retargeting; (2) Does pretraining lead to more efficient adaptation to new modalities/datasets compared to training from scratch; and (3) Does the shared diffusion backbone exhibit foundation-model-like properties, including data scaling and zero-shot composition of control signals?

\vspace{-1mm}
\subsection{Experiment Setup}

\paragraph{Evaluation Protocol.} We evaluate \method\ in two regimes: pretrained omni-modal motion generation and downstream finetuning. For pretraining, we consider language-, audio-, and human-reference-conditioned motion generation, where the model predicts future Unitree G1 whole-body trajectories from recent motion history and the corresponding control signal. For finetuning, we evaluate two downstream settings: text-conditioned generation on unseen data, and Pico keypoint-conditioned teleoperation. Unless otherwise specified, generated trajectories are executed by a pretrained HoloMotion~\cite{chen2026holomotion1} tracker. All evaluations use validation motions unseen during training. 

\paragraph{Metrics.} We evaluate \method\ along two axes: motion generation quality and tracking fidelity. For generation quality, we use modality-specific metrics, including Matching Score, R-Precision, FID, and Diversity~\cite{zhang2023generating} for language; BeatAlign, FID$_k$, FID$_g$, and PFC~\cite{li2021learn,tseng2023edge} for audio; and MPJPE, global MPJPE, end-effector error, velocity error, and acceleration error~\cite{jiang2022avatarposer,xie2026textop} for human-reference generation. For tracking fidelity, we report contact sliding, body jerk, tracker MPJPE, tracker global MPJPE, tracker velocity error and acceleration error, fall rate, and joint-limit violation rate~\cite{yi2022physical,cho2026safeflow}. For downstream finetuning, we additionally report task-specific metrics, including keypoint tracking error~\cite{nai2026humanoid} for Pico teleoperation. More details are provided in the supplementary material.

\subsection{Pretraining Experiments}
\input{tab/t2m_main}
\input{tab/a2m_main}

\input{tab/h2m_main}
\paragraph{Text to Motion.} We compare OMG against recent human motion generation models, including GENMO \cite{genmo2025}, HYMotion \cite{wen2025hy}, and Kimodo \cite{Kimodo2026}, where generated human motions are retargeted to the Unitree G1 when necessary via GMR \cite{ze2025gmr}. As shown in Table~\ref{tab:t2m_main}, \method\ achieves substantial gains in motion generation quality, while retaining high tracking fidelity, showcasing its capability to generate high-quality and robot-executable motions conditioned on text.

\paragraph{Audio to Motion.} We next evaluate audio-conditioned motion generation, comparing against both generalist and dedicated audio-to-motion baselines, including GENMO \cite{genmo2025}, LODGE \cite{li2024lodge}, and Bailando \cite{siyao2022bailando}. As shown in Table~\ref{tab:a2m_main}, \method\ maintains superior performance in audio alignment and motion quality, while achieving comparable performance in tracking fidelity. 
\paragraph{Human Reference to Motion.} Finally, we evaluate human-reference-conditioned motion generation. We compare against both optimization and learning based retargeting methods, including GMR \cite{ze2025gmr}, NMR \cite{zhao2026make}, PHC \cite{Luo2023PerpetualHC}, and OmniRetarget \cite{yang2025omniretarget}. As shown in Table~\ref{tab:h2m_main}, \method\ substantially outperforms these baselines in both motion quality and tracking fidelity, demonstrating that the learned generator can serve as an implicit yet highly-effective retargeting module for humanoid control.

\subsection{Finetuning Experiments}

\begin{table*}[t]
\centering

\begin{minipage}[t]{0.47\textwidth}
\centering
\caption{\textbf{Text-to-Motion Finetuning.}}
\scriptsize
\setlength{\tabcolsep}{5.5pt}
\renewcommand{\arraystretch}{1.05}
\resizebox{0.98\linewidth}{!}{
\begin{tabular}{lcccc}
\toprule
Finetuning Data \% & FID $\downarrow$ & R@1 $\uparrow$ & R@2 $\uparrow$ & R@3 $\uparrow$ \\
\midrule
\multicolumn{5}{c}{\textit{From Scratch}} \\
\midrule
\rowcolor{TableGray}
1\%
& 0.6615 & 0.0430 & 0.0762 & 0.1025 \\

10\%
& 0.1763 & 0.1162 & 0.2080 & 0.2998 \\

\rowcolor{TableGray}
100\%
& 0.1001 & 0.1689 & 0.2764 & 0.3779 \\

\midrule
\multicolumn{5}{c}{\textbf{\textcolor{darkred}{OMG}} \textit{Finetuned}} \\
\midrule
\rowcolor{TableGray}
1\%
& 0.1378 & 0.1533 & 0.2637 & 0.3428 \\

10\%
& 0.0886 & 0.1816 & 0.3096 & 0.3994 \\
\rowcolor{TableGray}
100\%
& \best{0.0827} & \best{0.2021} & \best{0.3115} & \best{0.4141} \\

\bottomrule
\end{tabular}
}
\label{tab:t2m_finetune}
\end{minipage}
\hfill
\begin{minipage}[t]{0.52\textwidth}
\centering
\caption{\textbf{Pico Keypoint-Based Teleoperation.}}
\scriptsize
\setlength{\tabcolsep}{5.5pt}
\renewcommand{\arraystretch}{1.05}
\resizebox{0.98\linewidth}{!}{
\begin{tabular}{lcccc}
\toprule
Finetuning Data \%
& MPJPE $\downarrow$
& g-MPJPE $\downarrow$
& E\_vel $\downarrow$
& E\_acc $\downarrow$ \\
\midrule
\multicolumn{5}{c}{\textit{From Scratch}} \\
\midrule
\rowcolor{TableGray}
1\%
& 128.24
& 448.09
& 24.34
& 17.13 \\

10\%
& 82.46
& 279.21
& 14.90
& 3.92 \\

\rowcolor{TableGray}
100\%
& 39.42
& 134.85
& 8.03
& 2.95 \\

\midrule
\multicolumn{5}{c}{\textbf{\textcolor{darkred}{OMG}} \textit{Finetuned}} \\
\midrule

\rowcolor{TableGray}
1\%
& 93.61
& 286.84
& 16.11
& 3.74 \\

10\%
& 52.99
& 164.91
& 9.91
& 3.09 \\

\rowcolor{TableGray}
100\%
& \best{25.73}
& \best{77.01}
& \best{5.70}
& \best{2.62} \\

\bottomrule
\end{tabular}
}
\label{tab:pico_finetune}
\end{minipage}
\end{table*}

\paragraph{Few-shot Text-to-Motion Finetuning.}
To evaluate whether motion-generation pretraining provides positive transfer to unseen data distributions, we finetune \method\ on AMASS-CMU, held out during pretraining. As shown in Table~\ref{tab:t2m_finetune}, the pretrained model consistently outperforms the from-scratch counterpart across all finetuning fractions. Notably, with only 1\% of the target data, finetuning already yields comparable motion quality to training from scratch with the full data budget.

\paragraph{Pico Keypoint-Based Teleoperation.}
To test if positive transfer extends to novel \textit{modalities}, we adapt \method\ to condition on Pico keypoints, enabling teleoperation. As shown in Table~\ref{tab:pico_finetune}, finetuned models consistently outperform training from scratch under the same data budgets, suggesting that the pretrained model serves as a reusable prior when incorporating a new control interface.


\subsection{Analysis}
\paragraph{Zero-shot Compositionality.} A hallmark of foundation models is \textit{compositional generalization}, i.e. the ability to combine capabilities learned independently. We test this by composing novel language and audio conditions at inference. As shown in Figure \ref{fig:exp_main}, the humanoid successfully follows both the language instruction and the audio rhythm simultaneously, producing motions that are qualitatively distinct from either condition applied alone.

\begin{wrapfigure}{r}{0.45\textwidth}
    \vspace{-1.5em}
    \centering
    \includegraphics[width=\linewidth]{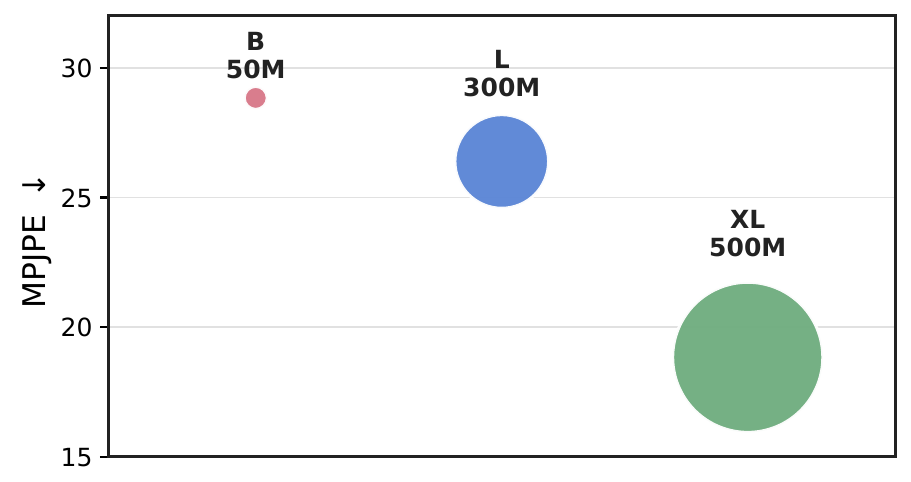}
    \caption{\textbf{Scaling \method\textbf{\textcolor{darkred}{-DiT}}.}}
    \label{fig:model_scaling}
    \vspace{-1em}
\end{wrapfigure}
\paragraph{Scaling Behavior.}
Finally, we ask whether motion generation is a scalable objective: do larger diffusion backbones yield better humanoid motion quality, given the same data and evaluation protocol?  To answer this, we pretrain three \method\textbf{\textcolor{darkred}{-DiT}} variants with increasing numbers of parameters. As shown in Figure~\ref{fig:model_scaling}, performance improves consistently with model size, suggesting that humanoid motion generation benefits from increased model capacity.

%% file: tab/t2m_main.tex
\begin{table}[t]
\caption{\textbf{Text-conditioned motion benchmark.} For better display, R@K, Fall, and J-Limit are reported in percentage. FID and C-Slide are scaled by $10^{-2}$ and $10^{-1}$, respectively. Lower is better for metrics marked with $\downarrow$, and higher is better for metrics marked with $\uparrow$.}
\centering
\scriptsize
\setlength{\tabcolsep}{2.8pt}
\renewcommand{\arraystretch}{1.08}
\resizebox{\textwidth}{!}{
\begin{tabular}{lcccccccccccccc}
\toprule
\multirow{2}{*}{Methods} 
& \multicolumn{8}{c}{Motion Generation Quality}
& \multicolumn{6}{c}{Tracking Fidelity and Execution Failures} \\
\cmidrule(lr){2-9} \cmidrule(lr){10-15}
& FID $\downarrow$
& Matching $\downarrow$
& R@1 $\uparrow$
& R@2 $\uparrow$
& R@3 $\uparrow$
& Diversity $\uparrow$
& C-Slide $\downarrow$
& Jerk $\downarrow$
& MPJPE $\downarrow$
& g-MPJPE $\downarrow$
& E\_vel $\downarrow$
& E\_acc $\downarrow$
& Fall $\downarrow$
& J-Limit $\downarrow$ \\
\midrule
\rowcolor{TableGray}
GENMO
& 43.78 & 1.33 & 12.99 & 23.83 & 30.86 & 1.28 & \best{2.66} & 173.54
& 50.35 & 150.71 & 6.91 & 2.51 & 1.46 & 29.10 \\

HYMotion
& 24.68 & 1.30 & 31.84 & 48.73 & 57.71 & 1.29 & 12.20 & 112.56
& 69.05 & 287.71 & 11.94 & 3.24 & 4.98 & 55.57 \\


\rowcolor{TableGray}
Kimodo-SMPLX-RP
& 38.44 & 1.31 & 26.27 & 39.36 & 46.48 & 1.25 & 4.52 & 189.39
& 43.51 & 198.96 & 7.80 & 2.64 & 2.64 & 23.63 \\

Kimodo-G1-RP
& 46.36 & 1.32 & 25.59 & 36.13 & 44.63 & 1.21 & 3.92 & 81.44
& \best{37.25} & 125.04 & 5.72 & 1.68 & 1.46 & 19.14 \\

\rowcolor{TableGray}
Kimodo-G1-SEED
& 50.22 & 1.33 & 22.46 & 33.69 & 41.11 & 1.20 & 4.30 & 99.47
& 37.26 & 161.63 & 6.38 & 1.84 & 1.56 & 20.70 \\

\midrule

\method\textbf{\textcolor{darkred}{-L}}
& 9.55 & 1.14 & 58.50 & 74.90 & 81.45 & \best{1.39} & 3.68 & 56.03
& 39.06 & \best{98.37} & \best{5.11} & \best{1.47} & \best{0.29} & \best{14.94} \\

\method\textbf{\textcolor{darkred}{-XL}}
& \best{6.03} & \best{1.11} & \best{65.43} & \best{80.37} & \best{86.91} & \best{1.39} & 3.98 & \best{52.31}
& 42.68 & 111.41 & 5.48 & 1.53 & 0.78 & 19.34 \\

\bottomrule
\end{tabular}
}
\label{tab:t2m_main}
\end{table}

%% file: tab/a2m_main.tex
\begin{table}[t]
\caption{\textbf{Audio-conditioned motion benchmark.} For better display, Fall and J-Limit are reported in percentage. Lower is better for metrics with $\downarrow$, and higher is better for metrics with $\uparrow$.}
\centering
\scriptsize
\setlength{\tabcolsep}{2.8pt}
\renewcommand{\arraystretch}{1.08}
\resizebox{\textwidth}{!}{
\begin{tabular}{lcccccccccccc}
\toprule
\multirow{2}{*}{Methods}
& \multicolumn{4}{c}{Audio Alignment and Target Motion Quality}
& \multicolumn{8}{c}{Target Physical Quality and Tracker Fidelity} \\
\cmidrule(lr){2-5} \cmidrule(lr){6-13}
& BeatAlign $\uparrow$
& FID$_k$ $\downarrow$
& FID$_g$ $\downarrow$
& PFC $\downarrow$
& C-Slide $\downarrow$
& Jerk $\downarrow$
& MPJPE $\downarrow$
& g-MPJPE $\downarrow$
& E\_vel $\downarrow$
& E\_acc $\downarrow$
& Fall $\downarrow$
& J-Limit $\downarrow$ \\
\midrule
\rowcolor{TableGray}
GENMO
& 0.57 & 506.07 & 9.42 & \best{0.02}
& \best{0.03} & \best{72.56} & \best{35.73} & 111.69 & \best{3.87} & \best{1.22} & 0.39 & 12.50 \\
LODGE
& \best{0.58} & 390.06 & 11.98 & 1.41
& 0.65 & 471.45 & 149.19 & 363.24 & 18.89 & 6.96 & 39.45 & 66.80 \\
\rowcolor{TableGray}
Bailando
& 0.56 & 134.90 & 15.71 & 0.39
& 0.25 & 117.45 & 38.30 & 106.39 & 5.55 & 1.99 & \best{0.00} & \best{1.76} \\
\midrule
\method\textbf{\textcolor{darkred}{-L}}
& 0.49 & 59.20 & 0.74 & 0.57
& 0.41 & 95.73 & 41.07 & \best{95.10} & 6.10 & 1.90 & \best{0.00} & 14.26 \\
\method\textbf{\textcolor{darkred}{-XL}}
& 0.51 & \best{40.46} & \best{0.63} & 0.64
& 0.43 & 106.06 & 41.47 & 97.60 & 6.03 & 1.94 & \best{0.00} & 15.43 \\
\bottomrule
\end{tabular}
}
\label{tab:a2m_main}
\end{table}

%% file: tab/h2m_main.tex
\begin{table}[t]
\caption{\textbf{Human-reference-conditioned motion benchmark.} For better display, Fall and J-Limit are reported in percentage. Lower / higher is better for metrics marked with $\downarrow$ / $\uparrow$.}
\centering
\scriptsize
\setlength{\tabcolsep}{2.8pt}
\renewcommand{\arraystretch}{1.08}
\resizebox{\textwidth}{!}{
\begin{tabular}{lccccccccccccc}
\toprule
\multirow{2}{*}{Methods}
& \multicolumn{7}{c}{Target Motion and Physical Quality}
& \multicolumn{6}{c}{Tracker-Executed Target Fidelity and Failures} \\
\cmidrule(lr){2-8} \cmidrule(lr){9-14}
& MPJPE $\downarrow$
& g-MPJPE $\downarrow$
& EE Error $\downarrow$
& E\_vel $\downarrow$
& E\_acc $\downarrow$
& C-Slide $\downarrow$
& Jerk $\downarrow$
& Target MPJPE $\downarrow$
& Target g-MPJPE $\downarrow$
& Target E\_vel $\downarrow$
& Target E\_acc $\downarrow$
& Fall $\downarrow$
& J-Limit $\downarrow$ \\
\midrule

\rowcolor{TableGray}
GMR
& 81.95 & 223.69 & 219.23 & 11.72 & 10.41 & 0.45 & 404.80
& 90.54 & 249.47 & 12.52 & 5.22 & 11.33 & 68.75 \\

NMR
& 173.77 & 390.82 & 477.08 & 14.87 & 7.43 & 0.77 & 97.75
& 102.28 & 299.27 & 11.26 & 3.32 & 14.84 & 87.89 \\

\rowcolor{TableGray}
PHC
& 51.44 & 153.44 & 152.18 & 5.06 & 3.97 & 0.45 & 131.38
& 47.80 & 115.92 & 5.86 & 1.99 & 3.71 & 39.45 \\

OmniRetarget
& 56.90 & 119.71 & 131.14 & 5.55 & 4.87 & \best{0.31} & 188.92
& 44.34 & 111.58 & 5.94 & 2.35 & 0.98 & 24.22 \\

\midrule

\rowcolor{TableGray}
\method\textbf{\textcolor{darkred}{-L}}
& 26.40 & 64.64 & 72.39 & 6.16 & 3.91 & 0.35 & \best{61.01}
& \best{42.81} & \best{110.66} & 5.63 & \best{1.62} & \best{0.20} & \best{22.66} \\

\rowcolor{TableGray}
\method\textbf{\textcolor{darkred}{-XL}}
& \best{18.84} & \best{47.83} & \best{52.26} & \best{5.03} & \best{3.69} & 0.32 & 64.04
& 43.07 & 112.60 & \best{5.55} & 1.63 & 0.39 & 23.05 \\

\bottomrule
\end{tabular}
}
\label{tab:h2m_main}
\end{table}

%% file: sections/conclusion.tex
\section{Conclusion}

We present \method, an omni-modal motion generation framework for generalist humanoid whole-body control. \method\ combines \method\textbf{\textcolor{darkred}{-Data}}, a large-scale corpus unified in Unitree G1 motion space, with \method\textbf{\textcolor{darkred}{-DiT}}, a shared diffusion backbone that maps language, audio, human references, and their compositions into robot-executable future motions. Experiments show strong motion quality, physical executability, sample-efficient adaptation, scaling behavior, and zero-shot compositionality, positioning scaling motion generation as a promising path toward humanoid foundation models.

%% file: sections/limitations.tex
\section{Limitations}

This work has several limitations. First, our training data mainly cover flat-ground motions; extending \method\ to uneven and in-the-wild terrain remains challenging. Second, \method\ uses a modular generator-tracker hierarchy, however, we focus only on motion generation for simplicity. We believe jointly adapting the generator and tracker, or incorporating execution feedback into generation, is an important direction for improving performance of the system, which we leave for future work.

%% file: app/theory.tex
\section{Extended Related Work}

\paragraph{Behavior Foundation Models for Humanoid Robots.}
Going beyond isolated skills, recent works have started to explore systems that capture broad, reusable behavioral knowledge for humanoid robots, often referred to as \emph{Behavior Foundation Models}~\citep{zeng2025behavior,yuan2025survey}. Using forward-backward representations, Meta Motivo \citep{tirinzoni2024metamotivo} and BFM-Zero \citep{li2025bfm} learn promptable latent-conditioned policies for zero-shot behavior selection, enabling multiple objectives such as motion tracking and goal reaching. Other representative works learn structured generative behavior priors with masked control interfaces for multiple low-level control modes~\citep{zeng2025behavior}, or introduce a generative middleware that rewrites reference commands for robust recovery~\citep{tao2026heracles}. Together, these works show that general humanoid behavior benefits from reusable priors, but their priors are primarily instantiated as motion primitives that operate close to low-level execution. In contrast, \method\ focuses on the upstream motion-generation interface, addressing how to translate heterogeneous \textit{human} commands, such as language, audio, or human reference motions, into robot-executable motion references.

%% file: app/data_details.tex
\section{Details on \method\textcolor{darkred}{-Data}}

\subsection{Dataset Composition}
\label{app:data_composition}

We report statistics for \method\textbf{\textcolor{darkred}{-Data}}, as shown in Table \ref{tab:data_composition}. All motion sequences are represented at a temporal resolution of 30 Hz. AMASS-CMU and Weizmann are held out during pretraining.

\begin{table}[ht]
    \centering
    \caption{\textbf{Statistics of the Pretraining Dataset by Conditioning Modality.} We curate publicly available motion datasets from graphics and humanoid domains and then apply careful filtering, annotation, and augmentation, yielding 1174.66\ hours of data in total. \labeltext, \labelaudio, and \labelhuman\ denote text descriptions, paired audio, and human reference motion, respectively.}
    \label{tab:data_composition}
    \resizebox{\linewidth}{!}{
    \begin{tabular}{llccc ccc}
        \toprule
        \multirow{2}{*}{Dataset} & \multirow{2}{*}{Label}
        & \multicolumn{3}{c}{Original}
        & \multicolumn{3}{c}{Processed} \\
        \cmidrule(lr){3-5} \cmidrule(lr){6-8}
        &
        & \# Samples & Avg. Length & Total Hours
        & \# Samples & Avg. Length & Total Hours \\
        \midrule
        AMASS \cite{mahmood2019amass}  & \labeltext, \labelhuman  & 17.9K & 1.4K & 94.8 & 51.1K & 111 & 52.7 \\
        AMASS (holdout CMU \& WEIZMANN) \cite{mahmood2019amass}  & \labeltext, \labelhuman  & 13.65K & 1.4K & 44.9 & 39.0K & 105 & 38 \\
        LAFAN1 \cite{harvey2020robust}  & \labeltext  & 40 & 6.6K & 2.5 & 977 & 240 & 2.2 \\
        OMOMO \cite{li2023object}  & \labeltext, \labelhuman  & 5.9K & 179 & 10 & 15K & 67 & 9.4 \\
        100style \cite{mason2022real}  & \labeltext, \labelhuman  & 0.8K & 2.9K & 22.12 & 10.7K & 223 & 22.10 \\
        HumanML \cite{fan2025zerozeroshotmotiongeneration}  & \labeltext, \labelhuman  & 26.8K & 219 & 54.5 & 25.7K & 218 & 52.04 \\
        Kungfu \cite{fan2025zerozeroshotmotiongeneration}  & \labelhuman  & 1.0K & 453 & 4.3 & -- & -- & -- \\
        MotionGV \cite{fan2025zerozeroshotmotiongeneration} & \labeltext, \labelhuman  & 56K & 140 & 727 & 53.8K & 128 & 637.14 \\
        fitness \cite{fan2025zerozeroshotmotiongeneration}  & \labeltext, \labelhuman  & 262 & 610 & 1.48 & 337 & 244 & 0.76 \\
        MotionLLAMA (subset) \cite{fan2025zerozeroshotmotiongeneration}  & \labeltext, \labelhuman  & 4.5K & 286 & 12 & 4.9K & 236 & 10.7 \\
        SnapMoGen \cite{snapmogen2025} & \labeltext, \labelhuman & 50K & 407 & 188.7 & 41.0K & 222 & 84.01 \\
        FineDance \cite{li2023finedance} & \labeltext, \labelaudio, \labelhuman & 0.2K & 4.0K & 14.6 & 6.0K & 240 & 13.26 \\
        BEAT2 \cite{liu2024emageunifiedholisticcospeech} & \labelaudio, \labelhuman & 1.7K & 3.4K & 60 & 417 & 2.0K & 8.06 \\
        AIST++ \cite{li2021learn} & \labeltext, \labelaudio, \labelhuman & 1.4K & 400 & 5.2 & 2.2K & 240 & 4.98 \\
        IDEA400 \cite{zhang2025motion,lin2023motionx} & \labeltext, \labelhuman & 12.5K & 176 & 20 & 10K & 177 & 17.24 \\
        PerMo \cite{kim2025personabooth} & \labeltext, \labelhuman & 6.6K & 140 & 8.6 & 6.5K & 139 & 8.38 \\
        BONES-SEED \cite{Kimodo2026} & \labeltext & 71K & 219 & 144 & 79K & 182 & 134.35 \\
        ChoreoMaster \cite{choreomaster2021} & \labeltext, \labelaudio & 2.7K & 46.7 & 1.2 & 2.5K & 47 & 1.11 \\
        CoMPAS3D \cite{burkanova2025salsanonverbalembodiedlanguage} & \labeltext, \labelaudio & 4.8K & 138 & 6.2 & 4.66K & 129 & 5.57 \\
        AIOZ-GDANCE \cite{aiozGdance} & \labeltext, \labelaudio & 6K & 1.1K & 60.8 & 5.8K & 1091 & 59.05 \\
        OpenDance \cite{zhang2025opendancemultimodalcontrollable3d} & \labeltext, \labelaudio & 5.0K & 300 & 14.0 & 4.9K & 300 & 13.61 \\
        \midrule
        \textbf{Total} & \labeltext, \labelaudio, \labelhuman
        & \textbf{288.8K} & \textbf{411} & \textbf{1496.9}
        & \textbf{364.5K} & \textbf{177} & \textbf{1174.66} \\
        \bottomrule
    \end{tabular}
    }
\end{table}

\subsection{Details on Retargeting}
\label{app:data_retargeting}

For motion represented in SMPL~\cite{SMPL} or SMPL-X~\cite{SMPLX} format, body parameters are directly translated and mapped onto the skeletal frame of the G1 robot via geometric optimization in GMR. For datasets derived from raw videos, we employ the GENMO framework~\cite{genmo2025} to recover 3D human body meshes, which are subsequently passed into GMR to compute G1 joint trajectories. For motions represented in the FBX format, we convert them using standardized joint-mapping heuristics or GMR, depending on the structural complexity of the source skeleton hierarchy.

\subsection{Details on Labeling}
\label{app:data_labeling}
We provide the prompt used for Seed-1.8 VLM annotation below:

\begin{promptbox}[label={prompt:vlm_design}]{Prompt used for VLM-based temporal motion annotation.}
You are a humanoid robot motion labeling model (VLM). Your task is to perform structured motion labeling for the input video, rather than generating descriptive text.

[Most Important Rule]
You must segment when there is a change in "motion atomic action". If the specific action changes, even if the overall style remains consistent, you still need to create a new segment.

[Core Requirement]
Please divide the input video into multiple consecutive time segments, and generate a structured action label for each segment.

The output of each segment must include:
1. start time
2. end time
3. style
4. action

[Segmentation Criteria]
You should decide whether to split mainly based on:
- whether the motion style changes
- whether a finer motion style / type can be judged to have changed
- whether the main action changes

If the motion remains continuous and no obvious change occurs, do not over-segment.

[Action Requirements]
1. action must be written in English
2. action should summarize the concrete motion performed in this segment
3. action should describe the motion itself as much as possible, and should not write overly abstract content
4. action should be concise, and no more than 25 words
5. action should be helpful for training a text-to-motion generation model

[Style Requirements]
1. style must be written in English
2. style should summarize the motion style of this segment, such as smooth, intense, rhythmic, explosive, etc.
3. style should be concise

[Output Format]
Please output in JSON format:
{
  "video_summary": "overall summary of the video",
  "segments": [
    {
      "start_time": 0.0,
      "end_time": 1.5,
      "style": "smooth",
      "action": "raise both arms and step slowly to the left"
    },
    {
      "start_time": 1.5,
      "end_time": 3.0,
      "style": "rhythmic",
      "action": "turn torso and swing arms in alternating beats"
    }
  ]
}

[Output Constraints]
1. Output valid JSON only
2. Do not output any explanatory text
3. Time must be in seconds
4. Keep all start_time and end_time values to one decimal place
5. Segments must be consecutive and non-overlapping
6. start_time must be less than end_time
7. video_summary should be a brief summary of the whole video

[Goal]
Your output should help train a motion generation model, so the labels should preserve motion content, motion style, and temporal changes as much as possible.
\end{promptbox}

\subsection{Segmentation Details}
\label{app:data_segmentation}

For text-conditioned sequences, motions are sliced according to the temporal boundaries of their language annotations. 
For audio-paired datasets, motion and audio sequences are segmented according to existing music phrase cuts or partitioned into uniform window lengths. 
For datasets with prohibitively long motion sequences, e.g., LAFAN1, 100style, FineDance, and AMASS, we further decompose each sequence into fixed-length sub-clips using a sliding-window strategy.

\subsection{Simulation-in-the-Loop Filtering Details}
\label{app:data_filtering}
For each generated motion, we first execute the sequence in the MuJoCo rigid-body simulator using the same tracker runtime used for deployment. The tracker predicts joint-space actions, which are clipped by \texttt{action\_clip}=10.0 and converted to desired joint positions. At each control step, we apply a joint-space PD executor for \texttt{control\_substeps}=10 simulation substeps:
\begin{equation}
    \tau = K_p (q_{\mathrm{des}} - q) - K_d \dot{q},
\end{equation}
where $q_{\mathrm{des}}$ is the desired joint configuration, $q$ and $\dot{q}$ are the simulated joint position and velocity, and $K_p, K_d$ are the actuator gains from the robot model.

After rollout, filtering is performed on the executed trajectory rather than on the original kinematic reference. We compute the root height $h_t$ and root tilt angle $\theta_t$ from the simulated root pose. A trajectory is rejected if non-finite states occur, or if a fall condition persists for at least 10 consecutive frames. Following the main text, a frame is considered a fall frame when
\begin{equation}
    h_t < 0.20
    \quad \mathrm{or} \quad
    \theta_t > 85^\circ
    \quad \mathrm{or} \quad
    (h_t < 0.35 \land \theta_t > 60^\circ).
\end{equation}
For datasets where strict robot feasibility is required, we additionally discard samples whose executed joint positions exceed the G1 joint limits. The remaining samples are kept as tracker-executed training data.

%% file: app/model_details.tex
\section{Details of \method\textbf{\textcolor{darkred}{-DiT}}}
\label{app:model_details}

\subsection{Motion Representation and Canonicalization}
\label{app:model_representation}

\paragraph{Motion Representation.} \method\textbf{\textcolor{darkred}{-DiT}} operates directly in the Unitree G1 motion space, without a learned motion tokenizer or latent autoencoder. Each motion frame is represented by a 125-dimensional feature vector
\begin{equation}
    \mathbf{x}_i =
    [\tilde{\mathbf{p}}_i,\tilde{\mathbf{r}}_i,\boldsymbol{\theta}_i,\tilde{\mathbf{j}}_i],
    \qquad \mathbf{x}_i \in \mathbb{R}^{125}.
\end{equation}
Here $\tilde{\mathbf{p}}_i \in \mathbb{R}^{3}$ is the root position in a canonical local coordinate frame, $\tilde{\mathbf{r}}_i \in \mathbb{R}^{6}$ is the root orientation represented by the continuous 6D rotation representation~\citep{zhou2019continuity}, $\boldsymbol{\theta}_i \in \mathbb{R}^{29}$ contains the G1 joint degrees of freedom, and $\tilde{\mathbf{j}}_i \in \mathbb{R}^{29 \times 3}$ contains local positions of non-root body links. This yields a $3 + 6 + 29 + 29 \times 3 = 125$-dimensional motion vector per frame.

\paragraph{Root Rotation Parameterization.} Concretely, a quaternion represents orientation as $\mathbf{q}=(q_w,q_x,q_y,q_z)\in\mathbb{R}^{4}$ with the unit-norm constraint $\|\mathbf{q}\|_2=1$. An axis-angle (rotation-vector) representation uses $\mathbf{r}=\alpha\mathbf{u}\in\mathbb{R}^{3}$, where $\mathbf{u}$ is a unit rotation axis and $\alpha$ is the rotation angle. The 6D representation stores the first two columns of a rotation matrix, $\mathbf{R}_{6D}=[\mathbf{r}_1,\mathbf{r}_2]\in\mathbb{R}^{6}$, and recovers a valid rotation by orthonormalizing these two vectors. We use 6D root rotation features by default. Compared with directly regressing quaternions, the 6D representation avoids unit-norm and sign discontinuities in the network output space; compared with axis-angle vectors, it provides a smooth over-parameterized representation that was empirically more stable during large-scale pretraining.

\paragraph{Canonicalization.} For every training sample, the model observes a history window of $L=10$ frames and predicts a future horizon of $H=60$ frames at 30 FPS. We use the \emph{last history frame} as the canonical anchor. Concretely, let $(\mathbf{p}_i, \mathbf{q}_i)$ denote the world-frame root position and quaternion orientation at motion frame $i$, and let $(\mathbf{p}_a, \mathbf{q}_a)$ be the anchor root state. We extract the yaw-only heading quaternion $\mathbf{h}_a$ from $\mathbf{q}_a$ and canonicalize the root trajectory as
\begin{equation}
    \tilde{\mathbf{p}}_i
    =
    \mathbf{h}_a^{-1} \otimes (\mathbf{p}_i - \mathbf{p}_a),
    \qquad
    \tilde{\mathbf{q}}_i
    =
    \mathbf{h}_a^{-1} \otimes \mathbf{q}_i,
\end{equation}
where $\otimes$ denotes quaternion rotation or composition depending on whether it is applied to a vector or a quaternion. The canonicalized quaternion $\tilde{\mathbf{q}}_i$ is then converted to the 6D root rotation feature $\tilde{\mathbf{r}}_i$. For each non-root body-link position $\mathbf{j}_{i,\ell}$, we similarly compute
\begin{equation}
    \tilde{\mathbf{j}}_{i,\ell}
    =
    \mathbf{h}_a^{-1} \otimes (\mathbf{j}_{i,\ell} - \mathbf{p}_a).
\end{equation}
The resulting canonical features remove arbitrary world-frame placement and heading, while retaining temporal motion trends from the history frames and the full G1 pose information required by the downstream tracker. At inference time, generated local features are decoded back to world-frame G1 states by composing them with the current anchor state. 

\begin{wrapfigure}{r}{0.52\textwidth}
\centering
\includegraphics[width=0.98\linewidth]{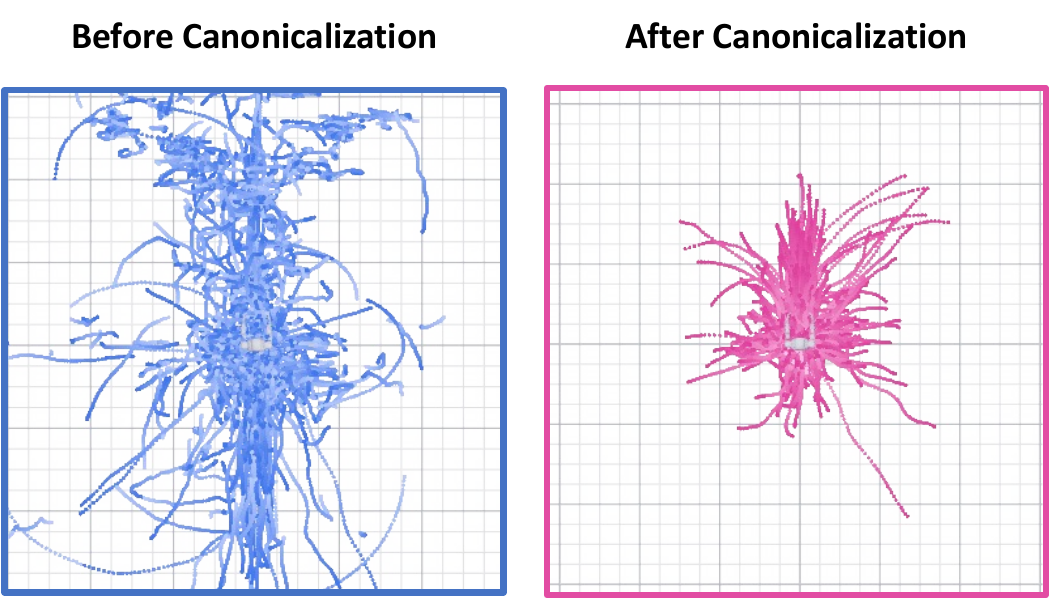}
\caption{\textbf{Canonicalization.}}
\label{fig:canonicalization}
\vspace{-1.0em}
\end{wrapfigure}

As illustrated in Figure~\ref{fig:canonicalization}, before canonicalization, root trajectories are widely scattered in the world frame: their coordinates are dominated by arbitrary global positions and headings, even when the underlying local motion patterns are similar. This creates a highly multi-modal input distribution that forces the model to explain nuisance variation unrelated to the commanded behavior. This motivates us to apply canonicalization. After canonicalization, trajectories collapse into a compact anchor-centered distribution, where remaining variation primarily reflects the robot's local motion dynamics rather than its absolute placement in the scene. This reduces the modeling burden of the diffusion backbone while preserving the temporal motion trend and full-body pose information required by the downstream tracker.

\paragraph{Normalization.} After canonicalization, every feature channel is normalized using training-set statistics:
\begin{equation}
    \bar{\mathbf{x}}_i
    =
    \frac{\mathbf{x}_i - \boldsymbol{\mu}}{\boldsymbol{\sigma}},
\end{equation}
where $\boldsymbol{\mu}$ and $\boldsymbol{\sigma}$ are computed from the training split. We clamp $\boldsymbol{\sigma}$ by a small positive minimum value for numerical stability. The diffusion model is trained and sampled in this normalized feature space, and outputs are denormalized before decoding to G1 joint states.

\subsection{Denoising Backbone}
\label{app:model_backbone}

The generator is implemented as a Diffusion Transformer trained with the $\mathbf{x}$-prediction objective. Here $\mathbf{x}$ denotes the normalized canonical motion feature defined above. Given a corrupted future motion segment $\mathbf{x}_{t+1:t+H}^{\tau}\in\mathbb{R}^{H\times125}$, diffusion timestep $\tau$, recent motion history $\mathbf{x}_{t-L:t}$, and a subset of available conditioning modalities $\mathcal{C}_s\subseteq\mathcal{C}$, the denoiser predicts the clean future motion as
\begin{equation}
\hat{\mathbf{x}}_{t+1:t+H}
=
\mathbf{x}_{\theta}
\left(
\mathbf{x}_{t+1:t+H}^{\tau},
\tau,
\mathbf{x}_{t-L:t},
\mathcal{C}_s
\right),
\qquad
\hat{\mathbf{x}}_{t+1:t+H}\in\mathbb{R}^{H\times125}.
\end{equation}

\paragraph{Architecture.} The denoiser first projects the corrupted future motion features into the Transformer hidden dimension. Each DiT block applies bidirectional temporal self-attention over the $H$ future frames. The recent history $\mathbf{x}_{t-L:t}$ and the condition subset $\mathcal{C}_s$ are encoded into context tokens and injected through cross-attention, while temporally aligned modalities such as audio or human reference motion can additionally be incorporated through frame-wise modulation. We use rotary positional embeddings (RoPE, \cite{su2024roformer}) for temporal self-attention, sinusoidal embeddings for the diffusion timestep $\tau$, and feed-forward blocks with GELU activations. The final prediction is a sequence of normalized clean motion features in the same 125D motion representation in Unitree G1 motion space. We list the architecture specifications for \method\textbf{\textcolor{darkred}{-DiT-B / L / XL}} in Table~\ref{tab:omg_dit_model_scales}. All variants share the same motion representation, diffusion objective, and conditioning interface architecture; they differ only in Transformer width, depth, and number of attention heads.

\begin{table}[t]
\centering
\caption{\textbf{Architecture variants of \method\textbf{\textcolor{darkred}{-DiT}}.}
To understand the scalability of motion generation, we introduce three variants of the denoising backbone randing from 50M to 500M parameters (B / L / XL), differing in transformer width, depth, and number of attention heads.}
\label{tab:omg_dit_model_scales}
\small
\setlength{\tabcolsep}{7pt}
\renewcommand{\arraystretch}{1.08}
\begin{tabular}{lcccc}
\hline
Model & Params & Hidden Dim. & Layers & Heads \tabularnewline
\hline
\method\textbf{\textcolor{darkred}{-DiT-B}}  & 50M  & 512  & 10 & 8  \tabularnewline
\method\textbf{\textcolor{darkred}{-DiT-L}}  & 300M & 1152 & 14 & 18 \tabularnewline
\method\textbf{\textcolor{darkred}{-DiT-XL}} & 500M & 1536 & 13 & 24 \tabularnewline
\hline
\end{tabular}
\end{table}

\paragraph{Denoising Objective and Sampling.} We train the model to predict the clean future motion $\mathbf{x}_0$ from a noisy future trajectory. We use discrete diffusion timesteps with a cosine noise schedule. Given a clean normalized future sequence $\mathbf{x}_0$, timestep $\tau$, and Gaussian noise $\boldsymbol{\epsilon}$, the noisy input is
\begin{equation}
    \mathbf{x}_{\tau}
    =
    \sqrt{\bar{\alpha}_{\tau}} \mathbf{x}_0
    +
    \sqrt{1-\bar{\alpha}_{\tau}} \boldsymbol{\epsilon}.
\end{equation}
At inference time, we use DDIM \cite{song2020denoising} sampling with 50 denoising steps and $\eta=0$. The default global guidance scale is 2.5 when a single guidance scale is used. For composed modalities, modality-specific scales can be set separately as described in Section~\ref{app:model_cfg}.

\subsection{Omni-Modal Condition Encoders}
\label{app:model_condition_encoders}

We provide details on condition encoders for different modalities in \method\textbf{\textcolor{darkred}{-DiT}}.

\paragraph{Motion History.}
The 10-frame motion history is first canonicalized and normalized using the same feature statistics as the future target. A two-layer MLP maps each history frame to the Transformer hidden dimension. The resulting history tokens are appended to the global context token sequence and interact with generated motion latents through per-block cross-attention layers.

\paragraph{Language.}
Text annotations are encoded by a frozen T5-Base \cite{raffel2020exploring} encoder with maximum length 50. The T5 language embeddings are projected into the transformer hidden dimension and injected as global context tokens alongside motion history through cross-attention. During training, text conditioning is randomly dropped with probability 0.3.

\paragraph{Audio.} Audio is encoded as frame-aligned 35D acoustic features. Given a raw waveform, we first average stereo channels to mono and peak-normalize the signal. For audio sample rate $s$ and motion frame rate $f=30$ Hz, the hop size is
\begin{equation}
    h=\mathrm{round}(s/f),
\end{equation}
and each audio frame uses a Hann window of length
\begin{equation}
    w=\max\{h,\mathrm{round}(0.05s)\},
\end{equation}
corresponding to at least a 50 ms analysis window. The feature vector at motion frame $i$ is extracted from the waveform segment starting at sample $ih$ and is defined as
\begin{equation}
    \mathbf{a}_i =
    \left[
    \mathbf{e}^{1:32}_i,\;
    \mathrm{rms}_i,\;
    \mathrm{zcr}_i,\;
    \mathrm{flux}_i+\mathrm{centroid}_i
    \right]\in\mathbb{R}^{35}.
\end{equation}
Here $\mathbf{e}^{1:32}_i$ are 32 log FFT band energies computed by splitting the magnitude spectrum into 32 bands and normalizing by the maximum band energy within the same frame, $\mathrm{rms}_i$ is root-mean-square energy, $\mathrm{zcr}_i$ is zero-crossing rate, $\mathrm{flux}_i$ is spectral flux from the previous audio frame, and $\mathrm{centroid}_i$ is the normalized spectral centroid. Audio features are trimmed or zero-padded to the motion horizon, with a validity mask for padded frames. During training, these features are stored offline as \texttt{.npy} arrays; at inference time, the same extractor can also be applied to raw WAV input before planning. The 35D features are passed through a LayerNorm and projected by an MLP into the denoiser hidden dimension. Since audio is temporally aligned with the future motion horizon, we inject it as a frame-wise condition using per-layer FiLM modulation. Audio conditioning is randomly dropped with probability 0.1 during training.

\paragraph{Human Reference Motion.}
We represent human reference motions as 66D frame-aligned features, corresponding to 22 human joints in 3D space. As with audio, the human reference signal is projected by an MLP and injected through per-layer FiLM modulation at the corresponding motion frame. Human-reference conditioning is randomly dropped with probability 0.5 during training.

For samples where a modality is unavailable, we set its availability mask to false. The mask gates the corresponding condition encoder and modulation, so the denoiser receives no conditioning signal from that modality. The same mechanism is used for modality dropout during training, allowing one shared model to train on heterogeneous datasets with different subsets of available conditions.

\subsection{Adapting to New Modalities}
\label{app:model_new_modalities}

\method\textbf{\textcolor{darkred}{-DiT}} is designed so that new control modalities can be included while preserving as much of the pretrained motion prior as possible. During adaptation, we reuse the pretrained denoising backbone and add new modalities with lightweight encoders.

For global conditioning signals, such as visual features for perceptive locomotion, the encoded tokens are appended to the cross-attention context, following the same interface as language and history tokens. For frame-aligned signals, such as Pico sparse keypoints for teleoperation, the encoded features are injected into each denoising block through lightweight adapters. In our implementation, these adapters can be per-layer FiLM modules or AdaLN-style modulation modules. The final linear layer of each newly added adapter is initialized to zero, so the adapter initially outputs no modulation and the pretrained generator's function is preserved at the start of finetuning.

This non-invasive initialization makes sample-efficient finetuning practical. The default adaptation setting trains the new modality encoder and its adapters while reusing the pretrained motion backbone. In our experiments, we use full finetuning across all backbone parameters, but parameter-efficient finetuning may also be feasible for simple tasks.

\subsection{Classifier-Free Guidance and Composition}
\label{app:model_cfg}

\paragraph{Classifier-Free Guidance and Multi-Modal Composition.} We train with modality dropout and use classifier-free guidance at inference time. For a single condition, the denoiser prediction is guided by comparing a conditional branch with a null-condition branch. For composed conditions, we additionally support modality-specific guidance branches. Given text, audio, and human-reference conditions, the guided prediction can be written as
\begin{equation}
    \hat{\mathbf{x}}_0 =
    \hat{\mathbf{x}}_0^{\varnothing}
    + \omega_{\mathrm{text}}
      (\hat{\mathbf{x}}_0^{\mathrm{text}} - \hat{\mathbf{x}}_0^{\varnothing})
    + \omega_{\mathrm{audio}}
      (\hat{\mathbf{x}}_0^{\mathrm{audio}} - \hat{\mathbf{x}}_0^{\varnothing})
    + \omega_{\mathrm{human}}
      (\hat{\mathbf{x}}_0^{\mathrm{human}} - \hat{\mathbf{x}}_0^{\varnothing}),
\end{equation}
where $\hat{\mathbf{x}}_0^{\varnothing}$ is the null-condition prediction and $\omega_{\mathrm{text}}$, $\omega_{\mathrm{audio}}$, and $\omega_{\mathrm{human}}$ are modality-specific guidance scales. By composing the guidance directions from different modalities, \method\textbf{\textcolor{darkred}{-DiT}} enables zero-shot composition of command combinations previously unseen during training.

\paragraph{Practical Implementation.} We instantiate classifier-free guidance in two ways during inference. If only a single global scale $\omega$ is specified, we use a full-condition branch:
\begin{equation}
    \hat{\mathbf{x}}_0
    =
    \hat{\mathbf{x}}_0^{\varnothing}
    +
    \omega(\hat{\mathbf{x}}_0^{\mathrm{all}} - \hat{\mathbf{x}}_0^{\varnothing}),
\end{equation}
where $\hat{\mathbf{x}}_0^{\mathrm{all}}$ is predicted with all available external modalities unmasked. If modality-specific scales are specified, we use separate guidance branches instead of a full-condition branch. 

Unless otherwise specified, we use global scale $\omega=2.5$ for single-modality motion generation. For composed audio-language generation, we use $\omega_{\mathrm{text}}=3.0$ and $\omega_{\mathrm{audio}}=1.5$, giving text a stronger semantic prior while retaining audio rhythm. For human-reference conditioning, we use $\omega_{\mathrm{human}}=2.0$. Intuitively, increasing $\omega_{\mathrm{text}}$ favors semantic instruction following, whereas increasing $\omega_{\mathrm{audio}}$ favors rhythmic and dance-style adherence. We provide further visualizations in Appendix \ref{app:visualize_cfg_comp}.


\paragraph{Extending Classifier-Free Guidance to New Modalities.} Classifier-free guidance extends naturally to newly added modalities. We construct a null branch and a modality-specific conditional branch for the new signal, and use
\begin{equation}
    \hat{\mathbf{x}}_0
    =
    \hat{\mathbf{x}}_0^{\varnothing}
    +
    \omega_{\mathrm{new}}
    (\hat{\mathbf{x}}_0^{\mathrm{new}} - \hat{\mathbf{x}}_0^{\varnothing}),
\end{equation}
where $\omega_{\mathrm{new}}$ is the guidance scale for the new modality. This branch can be combined with the existing modalities, enabling new control interfaces to be participate in compositional generation.

\subsection{Real-Time Deployment}
To enable real-time inference, we use a combination of existing runtime acceleration methods, including ONNX/TensorRT export, FP16 precision, and DiT caching \cite{liu2025timestep}. We provide runtime analysis and additional details in Appendix \ref{app:runtime_optimization_exp}. At deployment, the motion tracker runs onboard on the NVIDIA Orin chip, while the motion generator runs on an off-board NVIDIA RTX 4090 workstation connected to the robot through a wired Ethernet link.

\subsection{Training Hyperparameters}
\label{app:model_training}

We summarize key training hyperparameters for pretraining in Table~\ref{tab:omg_training_hparams}. We train in mixed bfloat16 precision with the AdamW optimizer \cite{loshchilov2017decoupled}, and apply standard techniques including gradient clipping, a linear-warmup cosine-decay learning-rate schedule, and weight decay. Training is completed on 8 NVIDIA A800 GPUs in under 10 hours.

\begin{table}[ht]
    \centering
    \caption{\textbf{Hyperparameters for \method\textbf{\textcolor{darkred}{-DiT}} Pretraining.} Unless otherwise specified, evaluation is conducted on the \textbf{\textcolor{darkred}{-L}} (300M) version of the model.}
    \label{tab:omg_training_hparams}
    \small
    \setlength{\tabcolsep}{6pt}
    \renewcommand{\arraystretch}{1.08}
    \begin{tabularx}{0.82\linewidth}{
        @{}>{\raggedright\arraybackslash}p{0.43\linewidth}
        >{\raggedright\arraybackslash}X@{}
    }
        \toprule
        \textbf{Hyperparameter} & \textbf{Value} \\
        \midrule

        \multicolumn{2}{@{}c}{\textbf{\textcolor{darkred}{Motion Setup}}} \\
        \midrule
        Prediction / history length & 60 / 10 frames \\
        Motion frame rate & 30 FPS \\
        Motion representation & 125D G1 features with 6D root rotation \\
        
        \addlinespace[0.35em]
        \midrule
        \multicolumn{2}{@{}c}{\textbf{\textcolor{darkred}{Optimization}}} \\
        \midrule
        Optimizer & AdamW \\
        Learning rate & $6.0 \times 10^{-5}$ \\
        Weight decay & $0.01$ \\
        LR schedule & Linear warmup, cosine decay \\
        Warmup steps & $2{,}000$ \\
        Minimum LR & $1.0 \times 10^{-6}$ \\
        Gradient clipping & Global norm $1.0$ \\

        \addlinespace[0.35em]
        \midrule
        \multicolumn{2}{@{}c}{\textbf{\textcolor{darkred}{Training}}} \\
        \midrule
        Max training steps & $100{,}000$ \\
        Precision & bf16 mixed precision \\
        GPUs & 8 \\
        Per-GPU batch size & 128 \\
        Global batch size & 1,024 \\

        \addlinespace[0.35em]
        \midrule
        \multicolumn{2}{@{}c}{\textbf{\textcolor{darkred}{Inference}}} \\
        \midrule
        Method & DDIM \\
        NFE calls & 50 \\
        \bottomrule
    \end{tabularx}
\end{table}

\clearpage




%% file: app/exp_details.tex
\section{Experiment Details}
\label{app:exp_details}

\subsection{Evaluation Protocols}
\label{app:exp_general_setup}

We provide details on evaluation protocols shared across different experiments, including the evaluated tasks, motion format, and data splits. Unless otherwise stated, all experiments follow these defaults.

\subsubsection{Evaluated Tasks and Motion Format}
\label{app:exp_tasks_motion_format}

\paragraph{Evaluated Tasks.} We evaluate six tasks: text-to-motion, audio-to-motion, human reference-to-motion, Pico keypoint-based teleoperation, text-to-motion finetuning, and perceptive locomotion. Although the conditions differ, all methods output Unitree G1 motion in the same G1 qpos space, making different methods and baselines directly comparable.

\paragraph{Motion Format.} Each robot frame is represented by a 36D qpos vector in G1 motion space:
\begin{equation}
    \mathbf{q}_t =
    \left[
    \mathbf{x}^{\text{root}}_t,\ 
    \mathbf{r}^{\text{root}}_t,\ 
    \mathbf{q}^{\text{joint}}_t
    \right]
    \in \mathbb{R}^{36},
\end{equation}
where $\mathbf{x}^{\text{root}}_t \in \mathbb{R}^{3}$, $\mathbf{r}^{\text{root}}_t \in \mathbb{R}^{4}$, and $\mathbf{q}^{\text{joint}}_t \in \mathbb{R}^{29}$ denote root position, root quaternion, and G1 joint DOFs. A sequence of length $T$ is $\mathbf{q}\in\mathbb{R}^{T\times36}$. For body-level metrics, we use forward kinematics:
\begin{equation}
    \mathbf{p}=\operatorname{FK}(\mathbf{q})\in\mathbb{R}^{T\times90},
\end{equation}
where the 90 dimensions are 3D positions of 30 G1 body/joint points. For baselines whose native outputs are not G1 qpos, such as SMPL-X, SMPL/SMPL-H motion, SMPL pickle files, or dance motion arrays, we first convert or retarget them to G1 qpos and then use the same evaluation pipeline.

\subsubsection{Evaluation Data Settings}
\label{app:exp_eval_data_sampling}

\paragraph{Evaluation Data Split.} Evaluations are performed on validation splits. The default generated sequence has 60 frames at 30 FPS. Generated motions are stored as G1 qpos, and body-level metrics are computed from FK-derived 90D body positions.

\begin{table}[ht]
    \centering
    \caption{\textbf{Evaluation data and sampling settings.}}
    \label{tab:exp_eval_settings}
    \resizebox{\linewidth}{!}{
    \begin{tabular}{llr}
        \toprule
        Task & Evaluation data & \# samples \\
        \midrule
        Text-to-motion & Caption-bearing validation samples & 1024 \\
        Audio-to-motion & Validation samples with audio/music features & 512 \\
        Human reference-to-motion & Validation samples with human reference motion & 512 \\
        Pico teleoperation & AMASS (CMU + WEIZMANN), with Pico keypoints & 512 \\
        Text-to-motion finetuning & AMASS (CMU + WEIZMANN), with captions & 1024 \\
        \bottomrule
    \end{tabular}
    }
\end{table}

As shown in Table \ref{tab:exp_eval_settings}, for text-to-motion, we sample 1024 caption-bearing validation conditions without replacement. R-precision uses batch size 32, giving 32 retrieval batches, and constructs candidates by dataset-stratified reordering. FID reference motions are sampled from real data; if the reference count is unspecified, it equals the generated sample count. AMASS CMU and WEIZMANN are used only for text-to-motion finetuning and Pico teleoperation from-scratch/finetuning experiments, and held out during pretraining, to avoid data leakage.

\subsection{Evaluation Metrics}
\label{app:exp_metrics}

In this section, we provide an overview of metrics used during evaluation, including general metrics shared across experiments and specific metrics for text-to-motion and audio-to-motion experiments. We additionally provide details on evaluator training for text-to-motion evaluation.

\subsubsection{Shared Evaluation Metrics}
\label{app:exp_physical_recon_metrics}
We use the same set of metrics for evaluating physical plausibility, reconstruction error relative to reference motion, and tracker execution failures across experiments, as shown in Table \ref{tab:exp_physical_recon_metrics}. Here $\mathbf{p}^{pred}$ and $\mathbf{p}^{ref}$ are predicted and reference G1 body positions, and $\tilde{\mathbf{p}}$ is the root-relative position. Velocity and acceleration are computed as first- and second-order finite differences of body positions.
\begin{equation}
    \mathbf{v}_{b,t,j}=\mathbf{p}_{b,t+1,j}-\mathbf{p}_{b,t,j},
    \quad
    \mathbf{a}_{b,t,j}=\mathbf{p}_{b,t+2,j}-2\mathbf{p}_{b,t+1,j}+\mathbf{p}_{b,t,j}.
\end{equation}
For c-slide, $\mathcal{C}$ is the set of valid foot-contact intervals and $\mathcal{S}_f$ contains sole proxy points for foot $f$. Fall Rate is determined by root height and tilt. For J-Limit, $e_{i,t,j}$ is the joint-limit violation magnitude and $\epsilon=10^{-4}$.

\begin{table}[ht]
    \centering
    \caption{\textbf{Shared Evaluation Metrics.}}
    \label{tab:exp_physical_recon_metrics}
    \resizebox{\linewidth}{!}{
    \begin{tabular}{lll}
        \toprule
        Metric & Formula & Description \\
        \midrule
        c-slide & $\frac{1}{|\mathcal{C}|}\sum_{(t,f)\in\mathcal{C}}\max_{p\in\mathcal{S}_f}\lVert \mathbf{s}^{xy}_{t+1,p}-\mathbf{s}^{xy}_{t,p}\rVert_2\cdot f_{\mathrm{fps}}$ & Horizontal foot sliding speed during contact \\
        Jerk & $\frac{1}{|\mathcal{V}|}\sum_{(t,j)\in\mathcal{V}}\lVert \mathbf{p}_{t+3,j}-3\mathbf{p}_{t+2,j}+3\mathbf{p}_{t+1,j}-\mathbf{p}_{t,j}\rVert_2\cdot f_{\mathrm{fps}}^3$ & Third-order motion variation \\
        g-MPJPE & $1000\cdot\frac{1}{BTJ}\sum_{b,t,j}\lVert \mathbf{p}^{pred}_{b,t,j}-\mathbf{p}^{ref}_{b,t,j}\rVert_2$ & Global body position error in mm \\
        MPJPE & $1000\cdot\frac{1}{BTJ}\sum_{b,t,j}\lVert \tilde{\mathbf{p}}^{pred}_{b,t,j}-\tilde{\mathbf{p}}^{ref}_{b,t,j}\rVert_2$ & Root-relative body position error in mm \\
        e-vel & $1000\cdot\frac{1}{B(T-1)J}\sum_{b,t,j}\lVert \mathbf{v}^{pred}_{b,t,j}-\mathbf{v}^{ref}_{b,t,j}\rVert_2$ & Velocity error in mm/frame \\
        e-acc & $1000\cdot\frac{1}{B(T-2)J}\sum_{b,t,j}\lVert \mathbf{a}^{pred}_{b,t,j}-\mathbf{a}^{ref}_{b,t,j}\rVert_2$ & Acceleration error in mm/frame$^2$ \\
        Fall Rate & $\frac{1}{N}\sum_{i=1}^{N}\mathbb{I}[\mathrm{sequence}_i\ \mathrm{fallen}]$ & Sequence-level fall rate \\
        J-Limit & $\frac{1}{N}\sum_{i=1}^{N}\mathbb{I}[\max_{t,j} e_{i,t,j}>\epsilon]$ & Sequence-level joint-limit violation rate \\
        \bottomrule
    \end{tabular}
    }
\end{table}

\subsubsection{Text-to-Motion Metrics}
\label{app:exp_t2m_metrics}

For text-to-motion, we additionally evaluate text-motion alignment and the generated motion distribution. Each generated G1 qpos is converted via FK to 90D body positions, encoded as a motion embedding, and compared with the corresponding text embedding. We report Matching Score, R-precision, FID, and Diversity, as shown in Table \ref{tab:exp_t2m_metrics}.

\begin{table}[ht]
    \centering
    \caption{\textbf{Text-to-motion metrics.}}
    \label{tab:exp_t2m_metrics}
    \resizebox{\linewidth}{!}{
    \begin{tabular}{lll}
        \toprule
        Metric & Formula & Description \\
        \midrule
        Matching Score & $\frac{1}{N}\sum_{i=1}^{N}\lVert \mathbf{m}_i-\mathbf{t}_i\rVert_2$ & Average distance between paired text-motion embeddings \\
        R-precision@K & $\frac{1}{N}\sum_{i=1}^{N}\mathbb{I}[\operatorname{rank}_i(i)\le K]$ & Whether the ground-truth text is in the top-$K$ retrieval results \\
        FID & $\lVert\boldsymbol{\mu}_r-\boldsymbol{\mu}_g\rVert_2^2+\operatorname{Tr}(\boldsymbol{\Sigma}_r+\boldsymbol{\Sigma}_g-2(\boldsymbol{\Sigma}_r\boldsymbol{\Sigma}_g)^{1/2})$ & Fr\'echet distance between real and generated motion embeddings \\
        Diversity & $\frac{1}{|\mathcal{P}|}\sum_{(a,b)\in\mathcal{P}}\lVert \mathbf{m}_a-\mathbf{m}_b\rVert_2$ & Average embedding distance between generated samples \\
        \bottomrule
    \end{tabular}
    }
\end{table}

Here $\mathbf{m}_i$ and $\mathbf{t}_i$ are motion and text embeddings, and $N$ is the number of samples. For R-precision, candidate texts are ranked by
\begin{equation}
    D_{ij}=\lVert \mathbf{m}_i-\mathbf{t}_j\rVert_2^2,
\end{equation}
and $\operatorname{rank}_i(i)$ is the rank of the paired text. We report $R@1$, $R@2$, and $R@3$. For FID, $(\boldsymbol{\mu}_r,\boldsymbol{\Sigma}_r)$ and $(\boldsymbol{\mu}_g,\boldsymbol{\Sigma}_g)$ are the empirical mean and covariance of real and generated motion embeddings. For Diversity, $\mathcal{P}$ contains randomly sampled generated embedding pairs; we use 300 pairs by default.

\subsubsection{Audio-to-Motion Metrics}
\label{app:exp_a2m_metrics}

Specific audio-to-motion metrics evaluate beat alignment and compare generated and real motions using kinetic and geometric statistics.

\begin{table}[ht]
    \centering
    \caption{\textbf{Audio-to-motion metrics.}}
    \label{tab:exp_a2m_metrics}
    \resizebox{\linewidth}{!}{
    \begin{tabular}{lll}
        \toprule
        Metric & Formula & Description \\
        \midrule
        Beat Align & $\frac{1}{|\mathcal{B}^a|}\sum_i \exp(-d_i^2/(2\sigma^2))$ & Temporal alignment between music and motion beats \\
        FID-k & $\lVert\boldsymbol{\mu}_{k,r}-\boldsymbol{\mu}_{k,g}\rVert_2^2+\operatorname{Tr}(\boldsymbol{\Sigma}_{k,r}+\boldsymbol{\Sigma}_{k,g}-2(\boldsymbol{\Sigma}_{k,r}\boldsymbol{\Sigma}_{k,g})^{1/2})$ & Fr\'echet distance over kinetic features \\
        FID-g & $\lVert\boldsymbol{\mu}_{geo,r}-\boldsymbol{\mu}_{geo,g}\rVert_2^2+\operatorname{Tr}(\boldsymbol{\Sigma}_{geo,r}+\boldsymbol{\Sigma}_{geo,g}-2(\boldsymbol{\Sigma}_{geo,r}\boldsymbol{\Sigma}_{geo,g})^{1/2})$ & Fr\'echet distance over geometric features \\
        PFC & $10000\cdot\frac{1}{T-2}\sum_t L_tR_t\hat{A}_t$ & Penalizes foot sliding during root upward acceleration \\
        \bottomrule
    \end{tabular}
    }
\end{table}

Here $\mathcal{B}^a$ is the music beat set and $d_i$ is the distance from music beat $i$ to the nearest motion beat. The Gaussian kernel width is $\sigma=3/f_{\mathrm{fps}}$, with $f_{\mathrm{fps}}=30$. FID-k uses kinetic features, FID-g uses geometric features, and PFC uses left/right foot sliding $L_t,R_t$ with normalized root upward acceleration $\hat{A}_t$.

\subsubsection{Details on Evaluator Training for Text-to-Motion Evaluation}
\label{app:exp_eval_encoders}

Text-to-motion embedding metrics use a separately trained text-motion evaluation encoder, following a CLIP-style contrastive retrieval protocol with symmetric InfoNCE loss. The evaluator is used only for evaluation and is not part of the generator. It contains a motion encoder and a text encoder that map motions and texts into a shared embedding space. We provide architecture details and key hyperparameters for evaluator training in Table \ref{tab:exp_motion_encoder_arch} and Table \ref{tab:exp_eval_encoder_hparams}, respectively.

\begin{table}[t]
    \centering
    \caption{\textbf{Motion Evaluator architecture used for evaluation.}}
    \label{tab:exp_motion_encoder_arch}
    \resizebox{\linewidth}{!}{
    \begin{tabular}{lll}
        \toprule
        Stage & Architecture & Output \\
        \midrule
        Movement encoder & Linear + LayerNorm + GELU + Dropout & Per-frame feature \\
        Hidden projection & Linear or Identity & Hidden sequence \\
        Temporal encoder & RoPE Transformer & Temporal feature sequence \\
        Pooling/projection & Mean pooling + MLP projection & Motion embedding \\
        Normalization & L2 normalization & Unit-norm embedding \\
        \bottomrule
    \end{tabular}
    }
\end{table}

\begin{table}[ht]
    \centering
    \caption{\textbf{Hyperparameters for Text-to-Motion Evaluator Training.}}
    \label{tab:exp_eval_encoder_hparams}
    \begin{tabular}{lc}
        \toprule
        Hyperparameter & Value \\
        \midrule
        Motion input dim & 90 \\
        Movement dim & 512 \\
        Hidden dim & 512 \\
        Output dim & 512 \\
        Transformer layers & 4 \\
        Attention heads & 8 \\
        MLP ratio & 4 \\
        \bottomrule
    \end{tabular}
\end{table}

The text encoder consists of a pretrained frozen T5-3B encoder followed by a trainable linear projection to the shared embedding space, where T5-3B denotes a T5 encoder with approximately 3 billion parameters. For sample $i$, the motion encoder takes the FK-derived 90D body-position sequence and the text encoder takes the corresponding text:
\begin{equation}
    \mathbf{m}_i = E_{\mathrm{motion}}(\mathbf{p}_i), 
    \quad
    \mathbf{t}_i = E_{\mathrm{text}}(\mathbf{c}_i),
\end{equation}
where $\mathbf{p}_i \in \mathbb{R}^{T\times90}$, $\mathbf{c}_i$ is the text condition, and $\mathbf{m}_i,\mathbf{t}_i$ are motion and text embeddings.

The two encoders are trained with symmetric InfoNCE:
\begin{equation}
    \mathcal{L}
    =
    \frac{1}{2}
    \left(
    \mathcal{L}_{m \rightarrow t}
    +
    \mathcal{L}_{t \rightarrow m}
    \right),
\end{equation}
where
\begin{equation}
    \mathcal{L}_{m \rightarrow t}
    =
    -\frac{1}{B}\sum_{i=1}^{B}
    \log
    \frac{
    \exp(\operatorname{sim}(\mathbf{m}_i,\mathbf{t}_i)/\tau)
    }{
    \sum_{j=1}^{B}
    \exp(\operatorname{sim}(\mathbf{m}_i,\mathbf{t}_j)/\tau)
    }.
\end{equation}
$\mathcal{L}_{t \rightarrow m}$ is symmetric. Here $B$ is the batch size, $\tau$ is temperature, and $\operatorname{sim}(\cdot,\cdot)$ is embedding similarity.

For standard text-to-motion evaluation, the encoder is trained on the full train split and evaluated on the validation split. For finetuning, it is trained and evaluated only on AMASS CMU and WEIZMANN, which are excluded from regular pretraining.

\subsection{Details of Evaluation Baselines}
\label{app:exp_baselines}

We compare against a wide variety of baselines from both the graphics and humanoid communities. For motion generation baselines in graphics, we use GMR \cite{joao2025gmr} to retarget the generated human poses into G1 motion space.

\subsubsection{Text-to-Motion Baselines}
\label{app:exp_t2m_baselines}

We compare against state-of-the-art motion generation baselines, including GENMO \cite{genmo2025}, HY-Motion \cite{wen2025hy}, and Kimodo \cite{Kimodo2026}. For Kimodo, we compare against variants trained with different datasets, including Bones Seed and Bones Rigplay, as well as generation in SMPL-X space or G1 space. All final outputs are evaluated as G1 \texttt{qpos\_36}. These baselines operate in a full-sequence manner, generating in advance instead of interactively in real time.

\begin{table}[ht]
    \centering
    \caption{\textbf{Text-to-motion Baselines.}}
    \label{tab:exp_t2m_baselines}
    \resizebox{\linewidth}{!}{
    \begin{tabular}{llll}
        \toprule
        Baseline & Original output space & Conversion to G1 & Final artifact \\
        \midrule
        GENMO~\cite{genmo2025} + GMR & SMPL-X motion & GMR retargeting & G1 \texttt{qpos\_36} \\
        HYMotion~\cite{wen2025hy} + GMR & SMPL/SMPL-H style motion & Convert to SMPL-X, then GMR & G1 \texttt{qpos\_36} \\
        Kimodo-SMPLX-RP-v1~\cite{Kimodo2026} + GMR & SMPL-X motion & GMR retargeting & G1 \texttt{qpos\_36} \\
        Kimodo-G1-RP-v1~\cite{Kimodo2026} & G1 qpos & Direct use & G1 \texttt{qpos\_36} \\
        Kimodo-G1-SEED-v1~\cite{Kimodo2026,seed2026seed1} & G1 qpos & Direct use & G1 \texttt{qpos\_36} \\
        \bottomrule
    \end{tabular}
    }
\end{table}

\subsubsection{Audio-to-Motion Baselines}
\label{app:exp_a2m_baselines}

For audio-to-motion baselines, we compare against motion generation models that can condition on multiple modalities, i.e. GENMO \cite{genmo2025}, as well as single-purpose state-of-the-art audio-to-motion generation baselines, such as LODGE~\cite{li2024lodge} and Bailando++~\cite{siyao2022bailando}. Their outputs are first converted to G1 qpos with GMR~\cite{joao2025gmr} and then evaluated with audio-motion and physical metrics. Specifically, GENMO~\cite{genmo2025} + GMR directly generates SMPL-X motion and retargets it to G1. LODGE~\cite{li2024lodge} and Bailando++~\cite{siyao2022bailando} are first converted to SMPL/SMPL-X-compatible representations and then retargeted. After conversion, all baselines share the same G1 qpos evaluation.

\begin{table}[ht]
    \centering
    \caption{\textbf{Audio-to-motion Baselines.}}
    \label{tab:exp_a2m_baselines}
    \resizebox{\linewidth}{!}{
    \begin{tabular}{lllll}
        \toprule
        Baseline & Condition input & Original output & Conversion to G1 & Final artifact \\
        \midrule
        GENMO~\cite{genmo2025} + GMR & Raw audio/music embedding & SMPL-X motion & GMR retargeting & G1 \texttt{qpos\_36} \\
        LODGE~\cite{li2024lodge} + GMR & Music feature \texttt{.npy} & Dance motion array & Convert to SMPL/SMPL-X, then GMR & G1 \texttt{qpos\_36} \\
        Bailando++~\cite{siyao2022bailando} + GMR & Raw audio/extracted feature & SMPL pickle & Convert to SMPL-X, then GMR & G1 \texttt{qpos\_36} \\
        \bottomrule
    \end{tabular}
    }
\end{table}

\subsubsection{Human Reference-to-Motion Baselines}
\label{app:exp_human_ref_baselines}

We represent the human reference using the 3D coordinates of 22 human joints, i.e.,
\begin{equation}
    \mathbf{h}\in\mathbb{R}^{T\times 22\times 3}.
\end{equation}
This is essentially a \emph{retargeting} problem, in which motion in human motion space is retargeted to G1 motion space. Here, we compare against well-established optimization-based retargeting methods, including GMR~\cite{joao2025gmr}, PHC~\cite{Luo2023PerpetualHC}, and OmniRetarget~\cite{yang2025omniretarget}, as well as the recently proposed learning-based method NMR~\cite{zhao2026make}.
\begin{table}[ht]
    \centering
    \caption{\textbf{Human reference-to-motion Baselines.}}
    \label{tab:exp_human_ref_baselines}
    \resizebox{\linewidth}{!}{
    \begin{tabular}{llll}
        \toprule
        Baseline & Type & Intermediate representation & Final artifact \\
        \midrule
        GMR~\cite{joao2025gmr} & Geometric/rule-based retargeting & Compact SMPL-X motion & G1 \texttt{qpos\_36} \\
        PHC~\cite{Luo2023PerpetualHC} & Physics/fitting-based retargeting & AMASS-style SMPL motion & G1 \texttt{qpos\_36} \\
        OmniRetarget~\cite{yang2025omniretarget} & Optimization-based retargeting & Holosoma-compatible SMPL-X motion & G1 \texttt{qpos\_36} \\
        NMR~\cite{zhao2026make} & Learned retargeting model & Standard SMPL-X motion & G1 \texttt{qpos\_36} \\
        \bottomrule
    \end{tabular}
    }
\end{table}

\subsection{Details on Finetuning and Scaling Experiments}
\label{app:exp_finetuning_model_scaling}

\paragraph{Datasets and Evaluation Splits.} We use the AMASS CMU and WEIZMANN subsets for text-to-motion finetuning and Pico teleoperation from-scratch/finetuning comparisons. To avoid leakage, these subsets are excluded from pretraining and used only for these experiments and evaluations. For scaling experiments, we pretrain on the exact \method\textbf{\textcolor{darkred}{-Data}} pretraining corpus. For sample-efficient finetuning with varying percentages of included data, data are controlled at the slice level. Each annotated motion segment is exhaustively sliced with a 2-second window at 30 FPS, so each slice has 60 frames and the full candidate slice pool is materialized. We shuffle this pool with a fixed random seed and keep the first specified percentage as the training set.

\paragraph{Setup.} We compare finetuning and from-scratch training under the same model size and split. Finetuning initializes from \method\textbf{\textcolor{darkred}{-DiT-L}}, the 300M pretrained checkpoint. The from-scratch setting uses the same architecture with randomly initialized parameters. For the model-parameter scaling-law experiment, we vary only the model size and keep all other hyperparameters unchanged.

We additionally include an adaptation experiment for perceptive locomotion in Appendix \ref{app:perc_loco}.

%% file: app/ext_exp.tex
\section{Extended Experiments and Visualizations}

We include additional experiments and visualizations in this section. We provide runtime acceleration analysis, extended visualizations on real-time omni-modal control, finetuning experiments on perceptive locomotion tasks, and analysis of classifier-free guidance.

\subsection{Runtime Acceleration Analysis}
\label{app:runtime_optimization_exp}

We report speedups from different inference optimization techniques for \method\textbf{\textcolor{darkred}{-DiT-B}} in Table~\ref{tab:latency_optimization}. Performance is measured on a text-conditioned 60-frame future chunk at 30 FPS, corresponding to a two-second planning horizon, on a single NVIDIA A800 GPU.
\begin{table}[ht]
    \centering
    \caption{\textbf{Runtime Acceleration Analysis.} Inference time is measured while generating a 60-frame future chunk with text conditioning, running \method\textbf{\textcolor{darkred}{-DiT-B}} on a single NVIDIA A800 GPU. We report the mean and standard deviation over five steady-state runs.}
    \label{tab:latency_optimization}
    \scriptsize
    \setlength{\tabcolsep}{3pt}
    \begin{tabular}{lrrlr}
        \toprule
        Optimization & Sampling (ms) & Denoiser infer (ms) & NFE (calls) & Speedup \\
        \midrule
        PyTorch eager & 1297.9 $\pm$ 12.5 & 1265.0 $\pm$ 12.4 & 100 (100) & 1.0$\times$ \\
        + \texttt{torch.compile} & 467.9 $\pm$ 2.9 & 435.6 $\pm$ 2.8 & 100 (100) & 2.8$\times$ \\
        + ONNX Runtime CUDA & 358.2 $\pm$ 24.2 & 337.0 $\pm$ 22.7 & 100 (50) & 3.6$\times$ \\
        + TensorRT engine & 127.3 $\pm$ 7.3 & 100.9 $\pm$ 4.3 & 100 (50) & 10.2$\times$ \\
        + TensorRT FP16 & 99.8 $\pm$ 5.2 & 73.3 $\pm$ 5.2 & 100 (50) & 13.0$\times$ \\
        + DiT cache & {51.7 $\pm$ 1.5} & {31.2 $\pm$ 1.3} & {38 (19)} & {25.1$\times$} \\
        \bottomrule
    \end{tabular}
\end{table}

Sampling time refers to the total time from receiving cached condition tensors to outputting future motion predictions, while Denoiser infer denotes the forward-pass time inside the sampler. NFE counts conditional/null CFG branch predictions; values in parentheses report backend forward calls after CFG parallelism.

\subsection{Extended Visualizations on Real-Time Omni-Modal Control}
\label{app:exp_visualization_setup}

\begin{figure}[p]
    \centering
    \includegraphics[width=\linewidth,height=0.93\textheight,keepaspectratio]{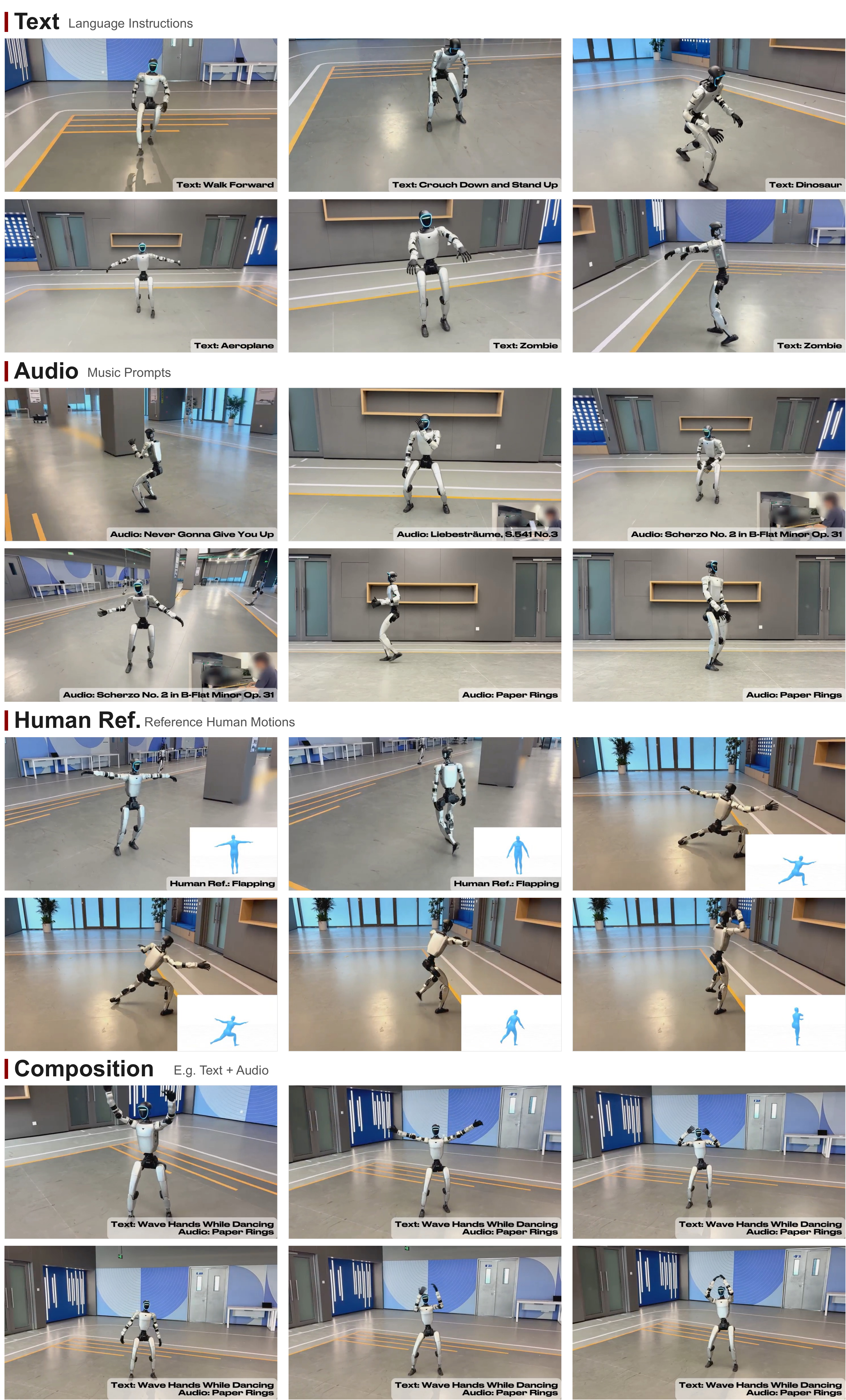}
    \caption{\textbf{Qualitative visualization.}
    Unitree G1 execution sequences produced by \method\ under text, audio, human-reference, and composed text-audio conditions. Frames are uniformly sampled within each sequence, and embedded prompts are preserved.}
    \label{fig:appendix_omnimodal_visualization}
\end{figure}

We provide extended qualitative visualizations on real-time omni-modal control. Motions are generated by the same pretrained model in real time, with conditions unseen during training. As shown in Figure \ref{fig:appendix_omnimodal_visualization}, \method\ generates diverse and robot-executable motions conditioned on various signals, showcasing strong capabilities as a foundation model for generalist humanoid control.

\clearpage

\subsection{Interactive Control with Temporal Composition}
\label{app:exp_interactive_control}

\begin{figure}[t]
    \centering
    \includegraphics[width=0.9\linewidth]{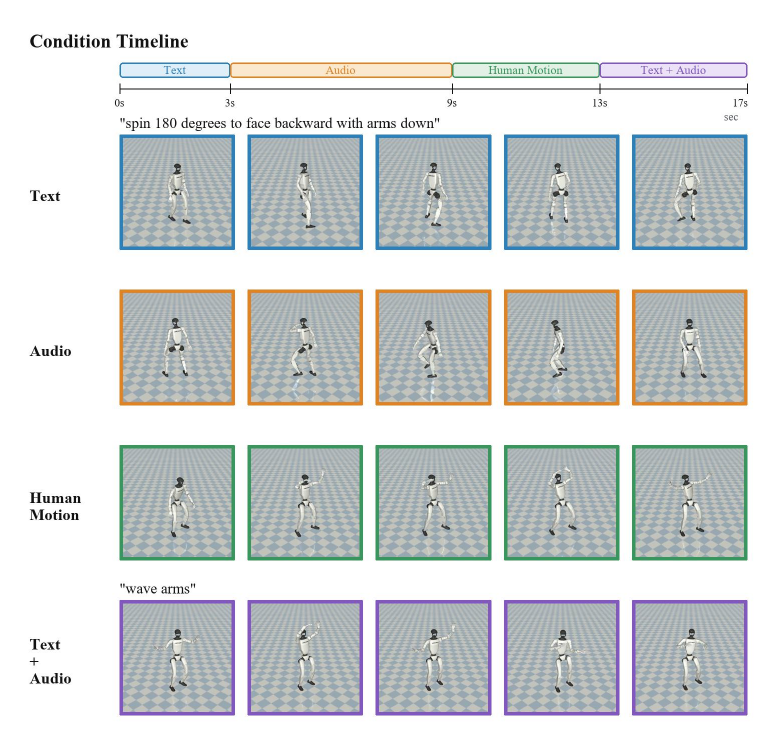}
    \caption{\textbf{Interactive Control: Composition in the Temporal Horizon.}}
    \label{fig:compositional_timeline_vertical_strip}
\end{figure}

We showcase our model's capability for real-time \emph{interactive} control. We feed the model time-varying commands from different modalities over the temporal horizon. As shown in Figure \ref{fig:compositional_timeline_vertical_strip}, \method\ follows local temporal conditions while maintaining smooth transitions between conditions.

\subsection{Analysis on Classifier-Free Guidance}
\label{app:visualize_cfg_comp}

In this section, we focus on answering the following questions:
\begin{enumerate}
    \item For single-modality conditioning, does classifier-free guidance improve instruction following? How does the scale of classifier-free guidance affect performance?
    \item For compositional modality conditioning, does increasing the guidance scale for one modality steer the behavior toward better alignment with instructions in that modality?
\end{enumerate}

\begin{figure}[t]
    \centering
    \includegraphics[width=\linewidth]{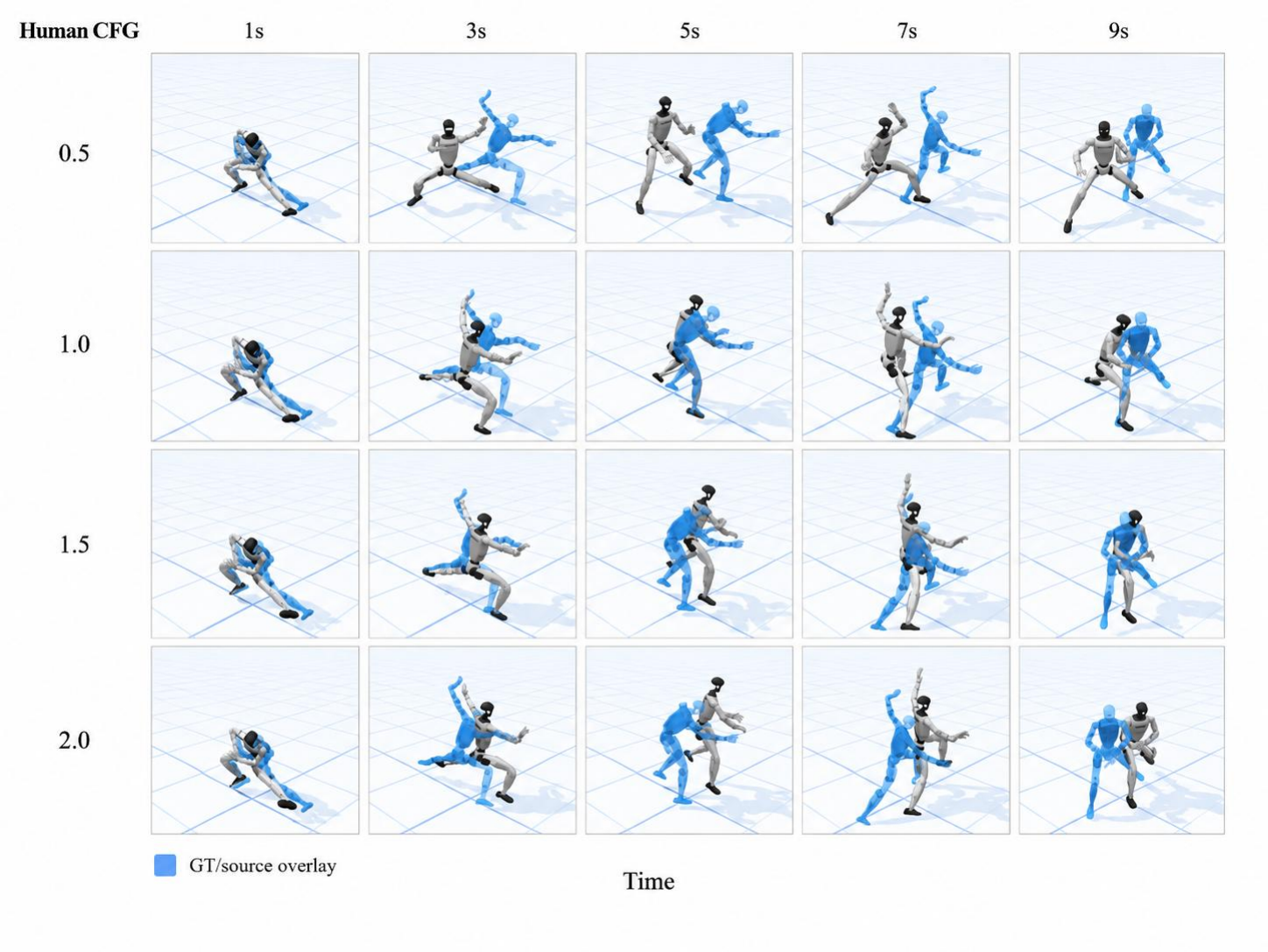}
    \caption{\textbf{Human-reference CFG sweep.} The translucent reference overlay shows the target motion, and the opaque robot shows the generated motion.}
    \label{fig:cfg_sweep_humanref}
\end{figure}

\paragraph{Single-Modality Classifier-Free Guidance.} We aim to understand whether classifier-free guidance improves the instruction-following capability of \method\ through the lens of human-reference conditioning. As shown in Figure~\ref{fig:cfg_sweep_humanref}, increasing the guidance scale for human-reference motion makes the generated motion more aligned with the reference, with the effect becoming more pronounced over longer horizons. We further measure MPJPE and g-MPJPE between the generated motion and ground truth under varying guidance scales. As shown in Table \ref{tab:cfg_sweep_humanref}, moderate guidance improves alignment, whereas overly strong guidance hurts performance.

\begin{table}[t]
    \centering
    \caption{\textbf{Human-reference CFG sweep.} We report MPJPE and g-MPJPE under varying classifier-free guidance scales for human-reference motion across different rollout lengths. MPJPE and g-MPJPE are reported in millimeters; lower is better.}
    \label{tab:cfg_sweep_humanref}
    \small
    \setlength{\tabcolsep}{6pt}
    \renewcommand{\arraystretch}{1.08}
    \begin{tabular}{lrrrrrrr}
        \toprule
        Human CFG & MPJPE & g-MPJPE & 1s & 3s & 5s & 7s & 9s \\
        \midrule
        0.5 & 190.3 & 557.0 & 62.7 & 407.2 & 236.1 & 213.2 & 91.5 \\
        1.0 & 96.6 & 261.8 & 42.9 & 59.3 & 117.3 & 144.0 & 216.6 \\
        1.5 & 74.0 & 218.9 & 38.5 & 51.6 & 108.9 & 93.2 & 139.2 \\
        2.0 & 71.0 & 343.9 & 38.3 & 60.4 & 109.5 & 82.6 & 73.4 \\
        \bottomrule
    \end{tabular}
\end{table}

\begin{figure}[t]
    \centering
    \includegraphics[width=\linewidth]{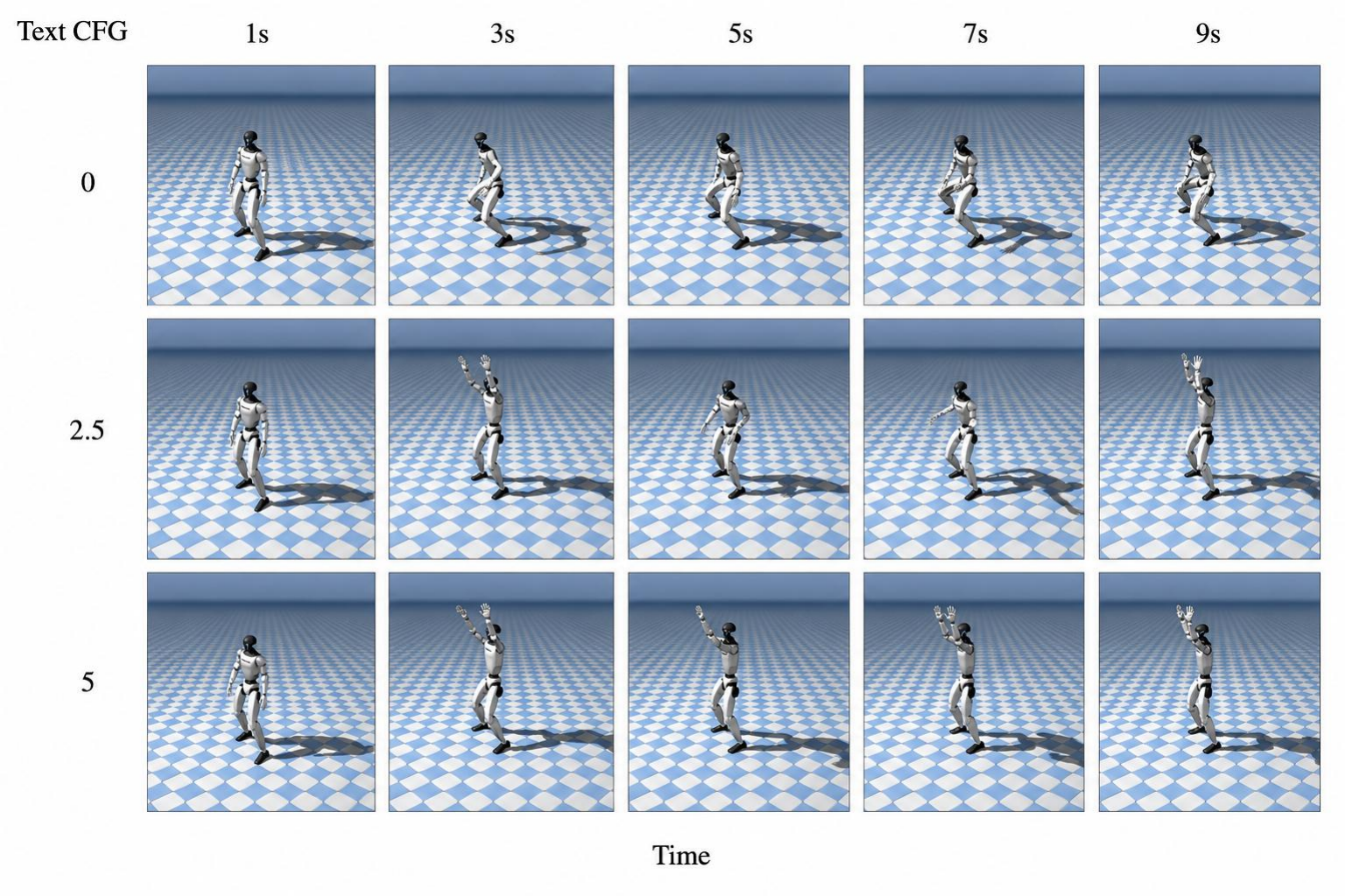}
    \caption{\textbf{Text-audio CFG sweep.} Columns show snapshots over time and rows vary the text guidance scale while keeping the audio condition fixed. Larger text guidance improves adherence to the language instruction in the composed audio-language setting.}
    \label{fig:cfg_sweep_composition}
\end{figure}

\paragraph{Multi-Modality Classifier-Free Guidance.} Next, we study the effects of guidance under multi-modal conditioning. As shown in Figure~\ref{fig:cfg_sweep_composition}, we compose text and audio conditions: the audio condition specifies the dance rhythm, while the language condition asks the robot to raise its hands. Increasing the text guidance scale strengthens semantic alignment to the language instruction while preserving the audio-conditioned temporal structure.

\subsection{Adaptation to New Modalities: Perceptive Locomotion}
\label{app:perc_loco}

\paragraph{Task and Environment.}
We evaluate whether the pretrained motion prior can be adapted to a new egocentric visual control task. We place three adjacent, differently colored square targets in front of the robot. The robot receives an egocentric RGB observations from a mounted camera and categorical target-color commands, and asked to locomote to the commanded square. 

\begin{figure}[t]
    \centering
    \begin{subfigure}[t]{0.22\linewidth}
        \centering
        \includegraphics[width=\linewidth]{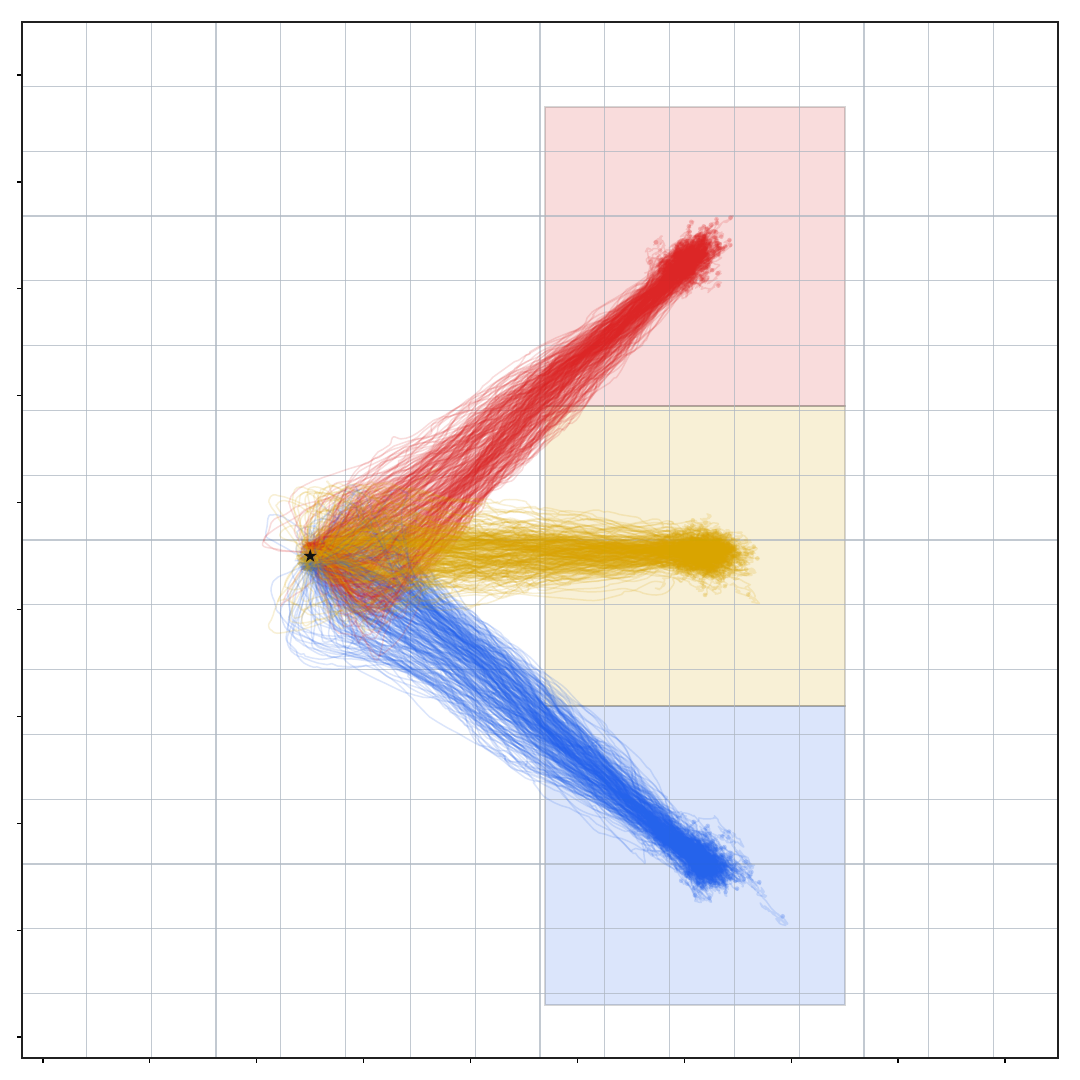}
        \caption{\textbf{Dataset.}}
    \end{subfigure}
    \hfill
    \begin{subfigure}[t]{0.39\linewidth}
        \centering
        \includegraphics[width=\linewidth]{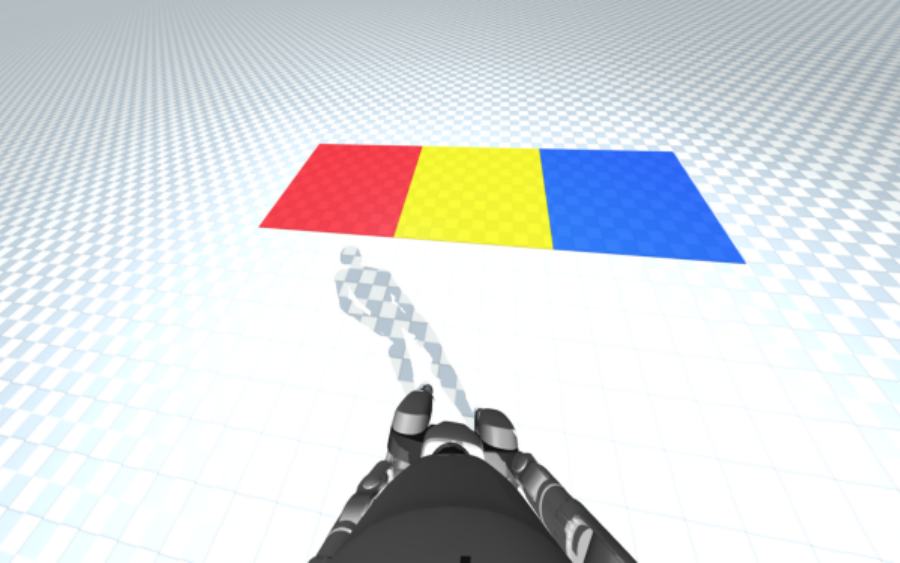}
        \caption{\textbf{Egocentric RGB input.}}
    \end{subfigure}
    \hfill
    \begin{subfigure}[t]{0.37\linewidth}
        \centering
        \includegraphics[width=\linewidth]{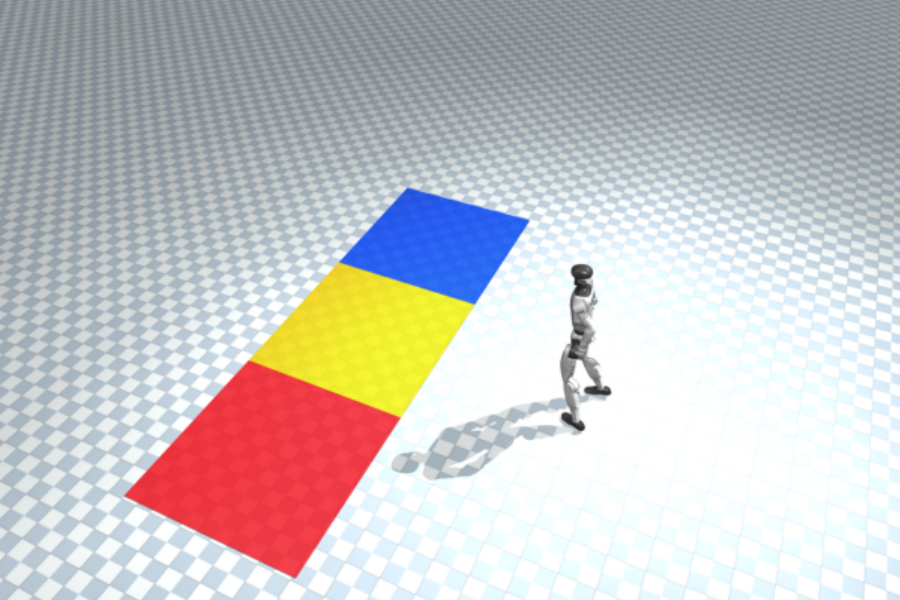}
        \caption{\textbf{Third-person layout.}}
    \end{subfigure}
    \caption{
    \textbf{Perceptive Locomotion setup.} The dataset contains 300 Kimodo-generated demonstrations per target color. The policy observes only low-resolution egocentric RGB and a discrete color command. The goal is to follow the command and locomote to the corresponding color.
    }
    \label{fig:perc_loco_setup}
\end{figure}

\paragraph{Training Data.}
We leverage Kimodo \cite{Kimodo2026} to generate demonstrations in advance, using its capability to condition on 2D paths. We collect 300 episodes per target color. Each episode contains 210 frames at 30 Hz, corresponding to roughly 7 s of source motion, and includes an appended target-hold segment so that the reference motion stops after reaching the target. To train the diffusion model, we sample 2 s windows using the same G1 state representation as the motion-generation model. The visual input is the current mounted-camera RGB frame resized to $64\times64$ and encoded as RGB patches. For fair comparison, we disable the language prompt and instead inject the target as a learned categorical color embedding for \{yellow, blue, red\}.

\paragraph{Model and Training Details.}
We compare two initializations under the same architecture and optimization hyperparameters.
The \emph{scratch} model is randomly initialized, while the \emph{pretrained} model initializes the motion backbone from the 300M mixed-modality \method\textbf{\textcolor{darkred}{-DiT-L}}. The newly introduced RGB patch encoder, visual cross-attention, and target-color embedding are trained in both cases. The visual patch gate is initialized to $0.05$ so that visual conditioning is active from the start of finetuning. Both models are trained on one GPU with a local batch size of 16, and a learning rate of $6\times10^{-5}$.

\paragraph{Evaluation and Results.} At test time, we use online replanning from a canonical standing G1 state.
Each replan samples a 2 s motion window, executes the first 0.5 s, and then replans from the updated state; the rollout budget is 210 source frames. We evaluate 30 held-out validation rollouts per target color, for 90 rollouts per checkpoint. Because this experiment is intended to measure whether the model can visually ground the target, we measure success as follows:
a rollout succeeds if, at some time in the trajectory, the root/pelvis horizontal position lies within the commanded target square during execution. As shown in Table \ref{tab:perc_loco_relaxed_results}, pretraining yields a higher success rate than training from scratch, showing that the pretrained motion prior transfers positively to new scenarios.

\begin{table}[t]
    \centering
    \caption{
    \textbf{Success Rate on Perceptive Locomotion.}
    We report success rate over 90 rollouts.
    }
    \label{tab:perc_loco_relaxed_results}
    \begin{tabular}{llcccc}
        \toprule
        Initialization & Step & Success Rate \\
        \midrule
        Scratch & 500 & 44/90 (48.9\%) \\
        Pretrained & 500 & \textbf{53/90 (58.9\%)} \\
        \bottomrule
    \end{tabular}
\end{table}

\begin{figure}[t]
    \centering
    \includegraphics[width=\linewidth]{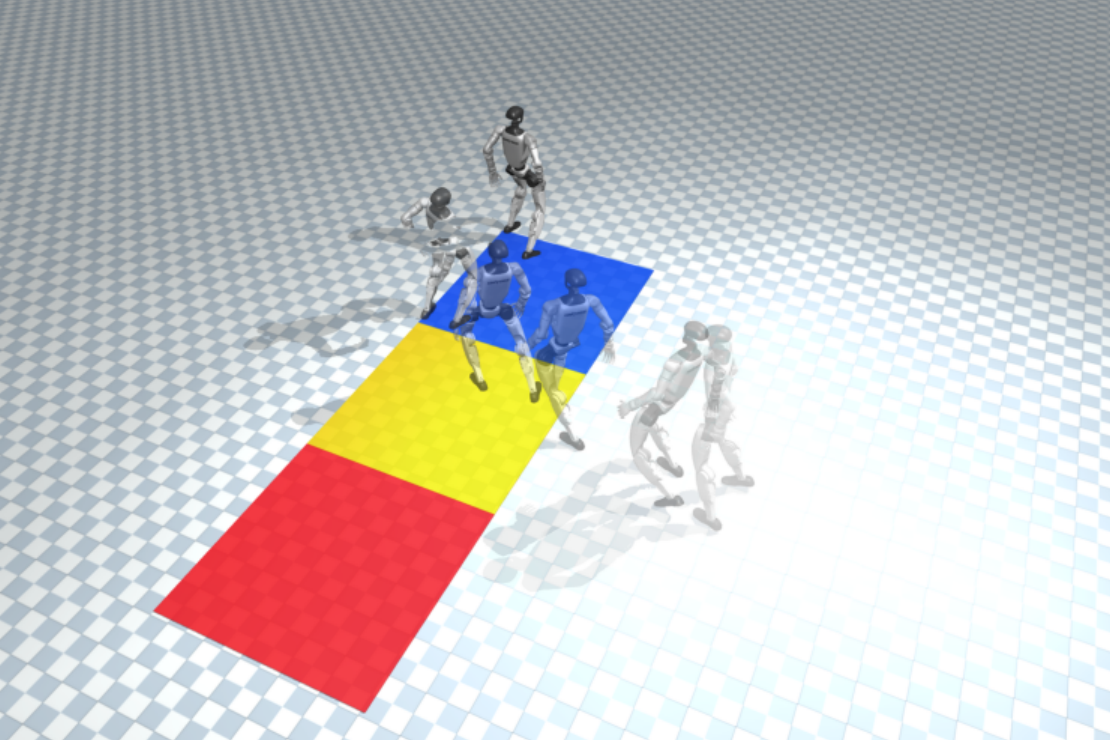}
    \caption{
    \textbf{Third-person timelapse of a successful rollout by the pretrained checkpoint.}
    The robot enters the commanded blue target.
    }
    \label{fig:perc_loco_success_timelapse}
\end{figure}